\pdfoutput=1

\documentclass[11pt]{article}

\usepackage{acl}

\usepackage{times}
\usepackage{latexsym}
\usepackage{graphicx} 
\usepackage{amsfonts, amsmath}
\usepackage{xcolor,colortbl}
\usepackage{multicol}
\usepackage{multirow}
\usepackage{tablefootnote}
\usepackage{longtable}
\usepackage{hyperref}
\usepackage{algorithm}
\usepackage{algorithmic}
\usepackage{multirow}
\usepackage{enumitem}
\usepackage{booktabs}
\usepackage{ulem}
\usepackage{soul}

\newcommand{\tabitem}{~~\llap{\textbullet}~~}

\DeclareMathOperator*{\argmax}{arg\,max}
\DeclareMathOperator*{\argmin}{arg\,min}
\definecolor{darkgray2}{rgb}{0.36, 0.36, 0.36}
\definecolor{LightCyan}{rgb}{0.8,0.9,0.8}
\definecolor{LightRed}{rgb}{1,0.75,0.75}
\definecolor{teal}{rgb}{0.98, 0.75, 0}
\definecolor{Gray}{gray}{0.93}
\definecolor{mintbg}{rgb}{.63,.79,.95}

\newcommand{\light}[1]{\textcolor{darkgray2}{#1}}

\newcommand\Tstrut{\rule{0pt}{2.6ex}}       
\newcommand\Bstrut{\rule[-0.9ex]{0pt}{0pt}} 
\newcommand{\TBstrut}{\Tstrut\Bstrut} 
\newcommand{\ST}{SemAE}
\newcommand{\SemAE}{Semantic Autoencoder}

\usepackage[T1]{fontenc}

\usepackage[utf8]{inputenc}

\usepackage{microtype}

\makeatletter
\newcommand{\ssymbol}[1]{^{\@fnsymbol{#1}}}

\usepackage[linecolor=orange]{todonotes}

\title{Unsupervised Extractive Opinion Summarization Using Sparse Coding}

\author{Somnath Basu Roy Chowdhury \qquad Chao Zhao \qquad Snigdha Chaturvedi \\ \texttt{\{somnath, zhaochao, snigdha\}@cs.unc.edu}\\ UNC Chapel Hill}

\begin{document}
\maketitle
\begin{abstract}
	Opinion summarization is the task of automatically generating  summaries that encapsulate information from
	multiple user reviews. We present {\textit{\SemAE}}  ({\ST}) to perform extractive opinion summarization in an unsupervised manner. {\ST} uses dictionary learning to implicitly capture semantic information from the review
	and  learns a latent representation of each sentence over semantic units. A semantic unit is supposed to capture an abstract semantic concept. Our extractive summarization algorithm leverages the {representations} to identify representative opinions among hundreds of reviews. {\ST} is also able to perform controllable summarization to generate aspect-specific summaries. 
	We report strong performance on \textsc{Space} and \textsc{Amazon} datasets, and perform experiments to investigate the functioning of our model. Our code is publicly available at \href{https://github.com/brcsomnath/SemAE}{https://github.com/brcsomnath/SemAE}.
\end{abstract}


\section{Introduction}
\label{sec:intro}
\textit{Opinion summarization} is the task of automatically generating digests 
for an \textit{entity} (e.g. a product, a hotel, a service, etc.), from user opinions in online forums. 
 Automatic opinion summaries  enable faster comparison, search, and better  consumer feedback understanding \cite{hu2004mining, pang2008lee, medhat2014sentiment}. 
 Although there has been significant progress towards  summarization 
\cite{rush2015neural, nallapati2016abstractive, cheng2016neural, see2017get, narayan2018ranking, liu2018generating}, existing approaches
rely on human-annotated reference summaries, which are scarce for opinion summarization. 
For opinion summarization, human annotators need to read hundreds of reviews per entity 
across different sources 
for writing a summary, which may not be feasible.

This lack of labeled training data  has prompted a series of works to leverage unsupervised or weakly-supervised techniques for opinion summarization \cite{mei2007topic, titov2008modeling, angelidis-lapata-2018-summarizing, angelidis2021extractive}. Recent works in this direction  have 
focused on performing opinion summarization in an abstractive setting \cite{coavoux2019unsupervised, isonuma2019unsupervised, bravzinskas2019unsupervised, amplayo2021unsupervised, iso-etal-2021-convex-aggregation, wang-wan-2021-transsum}. 
Abstractive models are able to produce fluent 
summaries using novel phrases. 
However, they 
suffer from problems common in text generation like hallucination \cite{rohrbach2018object}, text degeneration \cite{holtzman2019curious}, and topic drift \cite{sun2020improving}. 
Also, these approaches have been evaluated on small scales 
(10 reviews per entity or fewer), which does not reveal their utility in the real world  where there  are   hundreds of reviews per entity.

{To overcome these issues, another thread of works focuses on extractive opinion summarization, which creates summaries by selecting review sentences to reflect the popular opinions corresponding to an entity. A recently proposed extractive summarization approach is Quantized Transformer (QT) \cite{angelidis2021extractive}, which leverages vector quantization \cite{oord2017neural} for assigning texts to a latent representation that is supposed to capture a semantic sense. However, a text phrase can encapsulate multiple semantic senses, making this representation learning approach restrictive.}

{Building on the framework introduced by QT, we introduce an unsupervised extractive model, \textit{{\SemAE}} ({\ST}), which learns a representation of text over latent semantic units using \textit{dictionary learning} \cite{dumitrescu2018dictionary}. Similar to QT, {\ST} leverages Transformer \cite{vaswani2017attention} for sentence reconstruction to simultaneously learn latent semantic units and sentence representations. However, while QT assigns texts to a latent representation (codebook), {\ST} models text as a combination of semantics and forms a distribution over latent units (dictionary). This allows sentence representations to capture fine-grained and diverse semantics.  Unlike QT that relies on identification of aspect-specific head representations, we achieve controllable summarization by utilizing information-theoretic measures (such as relevance, redundancy, etc) on sentence representations. Our sentence selection algorithm is more flexible and allows a broader spectrum of controllable summarization. We experimentally show strong performance on two opinion summarization datasets. 
Our main contributions are:}

\begin{itemize}[topsep=1pt, leftmargin=*, noitemsep]
    \itemsep0mm
    \item We present {\SemAE} ({\ST}), which learns representation of sentences over latent semantic units. 
    \item We introduce novel inference algorithms for general and controllable summarization utilizing information-theoretic measures.
    \item We show that {\ST} outperforms previous methods using automatic and human evaluations.
    \item We perform 
    analysis to understand how the learnt 
    representations align with human semantics. 
\end{itemize}

\section{Related Work}  

{Unsupervised opinion summarization can be conducted either abstractively or extractively. Abstractive approaches aim to summarize the opinion text using novel phrases. Traditional statistical approaches create abstractive summaries using graphical paths \cite{ganesan2010opinosis} or hand-written templates \cite{di2014hybrid}. Recent neural approaches leverage the encoder-decoder architecture to aggregate information from multiple reviews and generate summaries accordingly \cite{chu2019meansum, bravzinskas2019unsupervised, iso-etal-2021-convex-aggregation, wang-wan-2021-transsum}.}


In contrast to abstractive approaches, extractive approaches rank and select a subset of salient sentences from reviews to form a concise summary \cite{kim2011comprehensive}. 
Saliency computation has been explored using traditional frequency-based approaches \cite{nenkova2005impact}, similarity with the centroid in the representation space \cite{radev2004centroid}, and lexical similarity with all sentences in a graph-based representation \cite{erkan2004lexrank}. Weakly supervised approaches \cite{angelidis-lapata-2018-summarizing, zhao2020weakly} extract opinions based on their aspect specificity, and nature of sentiment polarity. 

{Our work is most similar to the extractive opinion summarization QT~\cite{angelidis2021extractive} as discussed in Section~\ref{sec:intro}.}
It is also similar to neural topic model-based approaches \cite{iyyer2016feuding, he2017unsupervised, angelidis-lapata-2018-summarizing} that use a variant of dictionary learning \cite{elad2006image, olshausen1997sparse} to represent text as a combination of specific semantics (e.g. aspect, relationships etc). In contrast to these models, where text from same topics are trained to have similar representations using max-margin loss, {\ST} uses an autoencoder setup to capture diverse latent semantics.

\section{{Task Description}}
{We follow the task setup in \cite{angelidis2021extractive}, where given} a set of entities (e.g. hotels), 
a review set $\mathcal{R}_e = \{r_1, r_2, \ldots\}$ is provided for each entity $e$, where each review $r_i$ is a sequence of sentences $\{s_1, s_2, \ldots\}$. The review set $\mathcal{R}_e$ covers a range of aspects $\mathcal{A} = \{a_1, a_2, \ldots\}$ relating to the domain (e.g. \textit{service}, \textit{location} for hotels). We denote $S_e$ to be the set of sentences from all reviews for an entity $e$. {{\ST} is evaluated to perform two types of {extractive} opinion summarization  introduced by \citet{angelidis2021extractive}: 
(a) \textit{general summarization}, 
which involves selecting a subset of sentences  $O_e \subset S_e$ such that it best represents the reviews in $\mathcal{R}_e$, and (b) \textit{aspect summarization}, where the generated summary $O_e^{(a)} \subset S_e$ focuses  on a specific aspect $a \in \mathcal{A}$.}

\section{The Semantic Autoencoder}
\label{sec:training}
{The intuition behind {\SemAE} is that instead of representing text as a single latent semantic unit, we represent text as a distribution over latent semantic units using \textit{dictionary learning}.} 
{Learning semantic representations over a common dictionary makes them structurally aligned, enabling comparison of sentences using information-theoretic measures.} 

\begin{figure}[t!]
	\centering
 	\includegraphics[width=0.49\textwidth]{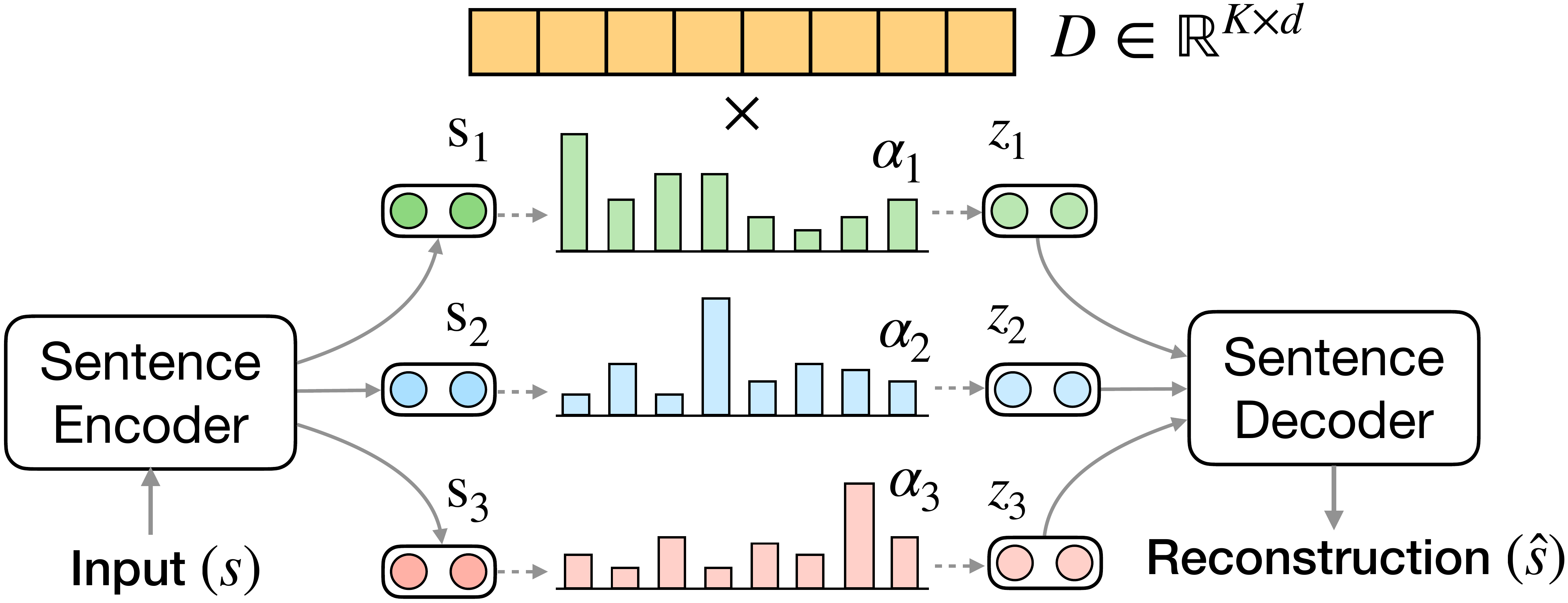}
	\caption{An example workflow of {\ST}. The encoder produces $H = 3$ representations ($\mathrm{s}_h$) for a review sentence $s$, which are used to generate latent representations over dictionary elements. The decoder reconstructs the input sentences using vectors ($z_h$) formed using latent representations ($\alpha_h$).}  
	\vspace{-8pt}
	\label{fig:ST}
\end{figure}

{{\SemAE} consists of three stages (i) \textit{sentence encoding} - an input sentence $s$ is converted into a multi-head representation ($H$ heads) using Transformer encoder $\{\mathrm{s}_h\}_{h=1}^H$; (ii) \textit{reconstruction} - a latent representation of head vectors $s_h$ is formed over elements of the dictionary $D \in \mathbb{R}^{K \times d}$, to produce {reconstructed} representations $\mathbf{z} = \{z_h\}_{h=1}^H$; 
and (iii) \textit{sentence decoding} - a Transformer-based decoder takes as input the reconstructed representations $\mathbf{z}$ 
to produce the output sentence $\hat{s}$. {\ST} is trained on the sentence reconstruction task.} 
The overall workflow of {\ST} is shown in Figure~\ref{fig:ST}. 


\subsection{Sentence Encoder}
\label{sec:encoder}
{We follow the setup of QT \cite{angelidis2021extractive} for sentence encoding. 
Each sentence $s$ starts with a special token \texttt{[SNT]}, which is fed to a Transformer-based encoder.} We only consider the final-layer representation of the \texttt{[SNT]} token $s_{\mathrm{snt}} \in \mathbb{R}^d$. The sentence representation $s_{\mathrm{snt}}$ is split into $H$ contiguous vectors $\{s_h'\}_{h=1}^H$, where $s_h' \in \mathbb{R}^{d/H}$. A multi-head representation is formed by passing $s_h'$ through a layer-normalization layer:
\begin{equation}
	\mathrm{s}_h = \mathrm{LN}(s_h'\mathrm{W}^T + b)
\end{equation}

\noindent where $\mathrm{W} \in \mathbb{R}^{d \times d/H}, b \in \mathbb{R}^d$ are trainable parameters and {$\mathrm{s}_h \in \mathbb{R}^d$ is the $h^{th}$ head representation}. 

\label{subsec:recon}

For each $s_h$, we obtain a latent representation $\alpha_h$ over the dictionary $D$, by reconstructing the encoded sentence representation $s_h$ as shown below
\begin{equation}
	z_h = \alpha_h D, \;\; \alpha_h = \mathrm{softmax}(s_h D^T) 
	\label{eqn:alpha_h}
\end{equation}

\noindent where the reconstructed vector $z_h \in \mathbb{R}^d$, and the latent representation $\alpha_h \in \mathbb{R}^K$. 
We hypothesize that the dictionary $D$ captures the representation of latent semantic units, and $\alpha_h$ captures the degree to which the text encapsulates a certain semantic.
The vectors formed $\mathbf{z}=\{z_h\}_{h=1}^H$ are forwarded to the decoder for sentence reconstruction. The dictionary $D$ and $s_h$ are updated simultaneously using backpropagation. {For summarization (Section~\ref{sec:inference}), different from QT, we consider $\alpha_h$ (not $z_h$) as the sentence representation.}


\subsection{Sentence Decoder}
{ 
{We employ a Transformer-based decoder that takes as input the reconstructed representations $\mathbf{z} = \{z_h\}_{h=1}^H$.} 
$\mathrm{MultiHead}(\mathbf{z}, \mathbf{z}, \mathbf{t})$ attention module in the decoder takes $\mathbf{z}$ as key and value, and the target tokens $\mathbf{t}$ as the query. The reconstructed sentence is generated from the decoder as $\hat{s} = \mathrm{Decoder}(\mathbf{z}, \mathbf{t})$. As our goal is sentence reconstruction, we set the target tokens to be same as the input sentence $s$.} {Prior work~\cite{angelidis2021extractive} has also used a similar Transformer-based decoder for sentence reconstruction but they attend directly over quantized head vector formed using codebook elements.}

{
A sentence can capture only a small number of semantic senses. We ensure this by enforcing sparsity constraints on the representations $\alpha_h$, so that $z_h$ is a combination of only a few semantic units.} The encoder, reconstructor and decoder are trained together to minimize the loss function:
\begin{equation}
	\mathcal{L} = \mathcal{L}_{\mathrm{CE}}(s, \hat{s}) + \lambda_1\sum\limits_{h}\lvert \alpha_h\rvert + \lambda_2\sum\limits_{h}H(\alpha_h)
\end{equation}

\noindent where $\mathcal{L}_{\mathrm{CE}}$ is the reconstruction cross-entropy loss of the decoder, and to ensure sparsity of $\alpha_h$ we penalize the L1-norm ($\lvert\alpha_h\rvert$) and its entropy $H(\alpha_h)$. 

\section{Summarization using {Latent} Representations}
\label{sec:inference}


We leverage the latent representations $\alpha_h$ generated by 
{\ST} 
to perform opinion summarization.\footnote{We experimented with different variations of the sentence selection scheme using $\alpha_h$ in Appendix~\ref{sec:variations}.}

\subsection{General Summarization}
For obtaining the general summary of an entity, we first compute a mean representation of all the review sentences in $S_e$, which represents the aggregate distribution over semantic units. Thereafter, the general summary is obtained as the collection of sentences that resemble the mean distribution.

Mathematically, every sentence $s$ is associated with a representation over dictionary elements $\alpha^s = [\alpha_1, \ldots, \alpha_H]$, where $\alpha^s \in \mathbb{R}^{H \times K}$. We form the mean representation of review sentences for an entity $S_e$ over dictionary elements as:
\begin{equation}
	\bar{\alpha} 
= \frac{1}{\lvert S_e \rvert}\sum_{s \in S_e} \alpha^s
\end{equation}

\noindent where 
$\alpha^s$ is the representation for sentence $s \in S_e$.

For general summarization, we compute the \textit{relevance} score $\mathcal{R}(\cdot)$ for each sentence $s$ based on its similarity with the mean representation $\bar{\alpha}$:
\begin{equation}
	\mathcal{R}(\alpha^s) = \Delta(\bar{\alpha}, \alpha^s) = - \sum_h \mathrm{KL}(\bar{\alpha}_h, \alpha_h^s)
	\label{eqn:relevance}
\end{equation}

\noindent where 
$\alpha^s_h$ is latent representation of sentence $s$ for the $h^{th}$ head. $\Delta(x, y)$ denotes the similarity between two representations $x$ and $y$. It is implemented as negation of the sum of KL-divergence between head representations. 
We also experimented with other divergence metrics and observed similar summarization performance (Appendix~\ref{sec:ext-analysis}). 

\begin{figure}[t!]
	\centering
	\includegraphics[width=0.48\textwidth]{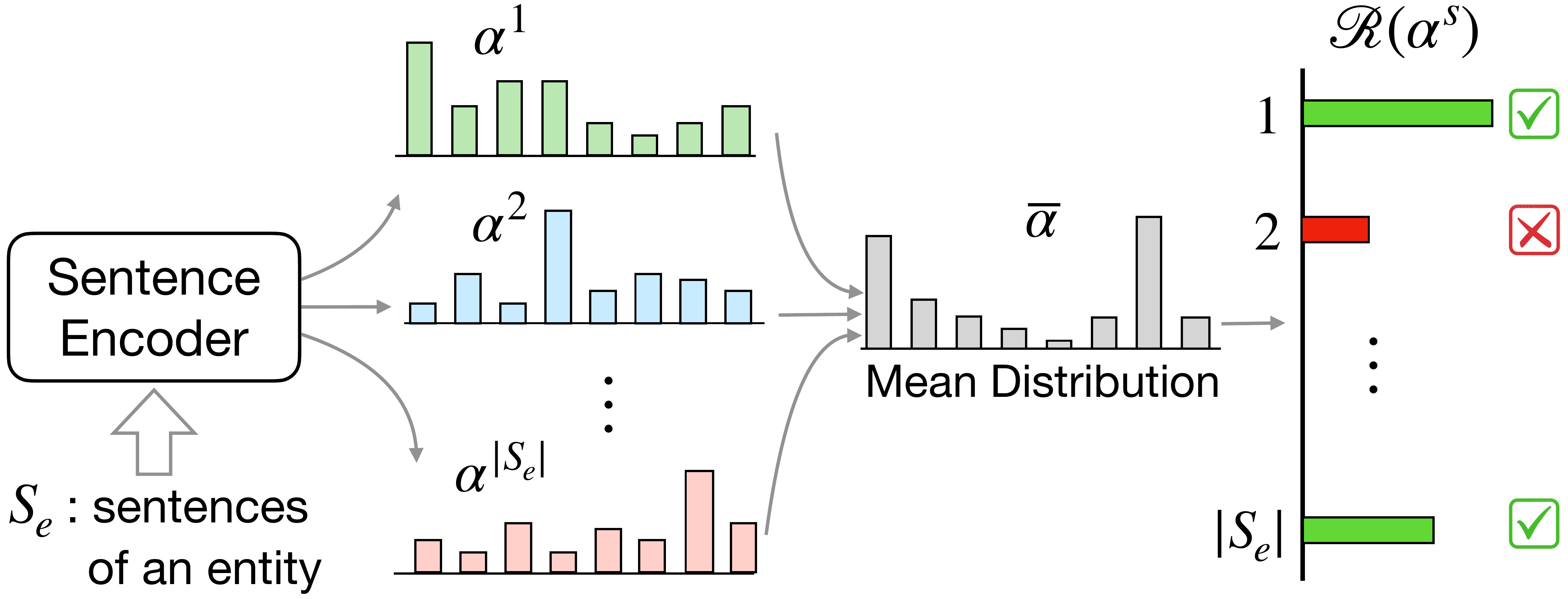}
	\caption{ General summary generation routine. The relevance score of each sentence w.r.t mean representation is computed, and top $N$ sentences ($O_e$) with highest $\mathcal{R}(\cdot)$ are selected as the summary.}
	\label{fig:Inference}
	\vspace{-8pt}
\end{figure}

We rank sentences according to descending order of  $\mathcal{R}(\cdot)$ and select the top $N$ (a constant hyperparameter, $N < \lvert S_e\rvert $) sentences  as the summary $O_e$ (shown in Figure~\ref{fig:Inference}). 
The extracted summary is a concatenation of the text from $N$ selected input sentences (Input ($s$) in Figure~\ref{fig:ST}).
However, modeling relevance only using $\Delta(\cdot, \cdot)$ results in selection of similar sentences. We overcome this by designing variations of our system that have additional information-theoretic constraints.

\noindent (a) \textbf{Redundancy}: 
	We introduce diversity in the generated summary  by penalizing sentences that have a high similarity value with already selected sentences. 
	This is achieved by adding the \textit{redundancy} term in relevance score:
	\begin{equation}
		\mathcal{R}(\alpha^s, \hat{O}_e) = \Delta(\bar{\alpha}, \alpha^s) - \gamma \max_{s' \in \hat{O}_e}\Delta({\alpha}^{s'}, \alpha^s)
	\label{eqn:red}
	\end{equation}
	where $\hat{O}_e$ is the set of  sentences selected so far for the summary. The selection routine proceeds in a greedy fashion by choosing $s_0 = \argmax_{s \in S_e} \Delta(\bar{\alpha}, \alpha^s)$ when $\hat{O}_e = \phi$.

	\noindent(b) \textbf{Aspect-awareness}: 
	Another drawback with sentence selection using $\Delta(\cdot, \cdot)$ is that the summary frequently switches context among different aspects (example shown   in Table~\ref{tab:ablated-summaries}).   
	To mitigate this issue, we identify the aspect of a review sentence {using occurrences of aspect-denoting 
	keywords provided in the dataset (Section~\ref{sec:aspect-summ}).}
	{We then cluster the sentences into aspect-specific buckets  $\{S_e^{(a_1)}, S_e^{(a_2)}, \ldots\}$ and rank sentences within each bucket. } 
	We ignore sentences that are not part of any bucket. We select sentences using two different strategies: 
	
	\begin{itemize}[leftmargin=*, topsep=0pt]
	\itemsep0mm
	    \item We iterate over  sentence buckets $\{S_e^{(a_i)}\}$ and select the first $m$ sentences ranked according to $\mathcal{R}(\alpha^s)$, from each bucket.
	    \item 
	    We prevent selection of similar sentences from a bucket by introducing the redundancy term.
	    We iterate over individual buckets and select first $m$ sentences ranked according to their relevance $\mathcal{R}(\alpha^s, \hat{O}_e^{(a)})$ (Equation~\ref{eqn:red}).
	\end{itemize}

\subsection{Aspect Summarization}
\label{sec:aspect-summ}
{\ST} can perform aspect summarization without needing additional training. For this, we require a small set of 
keywords 
to identify sentences that talk about an aspect. For example, \textit{food} aspect is captured using keywords: ``breakfast'', ``buffet'' etc.

For a given aspect $a$, let the keyword set be $Q_a = \{w_1, w_2, \ldots \}$. We use $Q_a$ to identify a set of sentences $S_e^{(a)}$ for each entity $e$, belonging to aspect $a$  
from a held-out dev set $S_{dev}$. 
Similar to general summarization, we proceed by computing the mean representation of sentences $S_e^{(a)}$ belonging to the aspect  $a$:
\begin{equation}
	\bar{\alpha}^{(a)} 
	=  \frac{1}{\lvert S_e^{(a)}\rvert} \sum_{s \in S_e^{(a)}} \alpha_s
\end{equation}

{We then select sentences most similar to the mean representation as the summary.}


\noindent(a) \textbf{Informativeness}: Sentences selected for aspect summarization should talk about the aspect but not the general information. We model \textit{informativeness} \cite{peyrard2018simple} by ensuring that a selected sentence representation $\alpha_s$ resembles the aspect mean $\bar{\alpha}^{(a)}$, but is divergent from the overall  representation mean $\bar{\alpha}$, for a given entity $e$. For an aspect $a$, we iterate over sentences in $S_e^{(a)}$ and compute the relevance score for a sentence $s$ as follows:
\begin{equation}
\mathcal{R}_a(\alpha^s) = \Delta(\bar{\alpha}^{(a)}, \alpha^s) - \beta \Delta(\bar{\alpha}, \alpha^s)
\label{eqn:aspect-inf}
\end{equation}

We rank sentences $s \in S_e$ according to their \textit{aspect-specific relevance} score $\mathcal{R}_a(\cdot)$, and select first $N$ sentences as the summary for aspect $O_e^{(a)}$.\footnote{We experimented with incorporating the informativeness term in general summarization also but did not find it useful (see Appendix~\ref{sec:inf} for more details).}

\section{Experimental Setup}
{In this section, we discuss the experimental setup, results and analysis.}

\subsection{Datasets}
We evaluated our model on two public customer review datasets \textsc{Space} hotel reviews \cite{angelidis2021extractive} and \textsc{Amazon} product reviews \cite{he2016ups, bravzinskas2019unsupervised}. 
The dataset statistics are reported in Table~\ref{tab:stat}. 
Test sets of both datasets contain three human-written general summaries per entity. {The \textsc{Space} corpus was created in a two-step process of sentence selection and then summarization of selected sentences by annotators (further details in Appendix~\ref{sec:dataset}).} \textsc{Space} dataset also provides human-written summaries for six different aspects of hotels: \textit{building}, \textit{cleanliness}, \textit{food}, \textit{location}, \textit{rooms}, and \textit{service}.

\subsection{Implementation details}
{We build on the implementation framework introduced by \citet{angelidis2021extractive} for our experiments}. We used a 3-layer Transformer with 4 attention heads as the encoder and decoder. The input and hidden dimensions are 320.  
The encoder and decoder for {\ST} was trained for 4 warmup epochs, before the dictionary learning based reconstruction component was introduced. We split the encoded vector into $H=8$ head representations. We have $K=1024$ dictionary elements, each with dimension $d=320$. The dictionary elements are initialized using $k$-means clustering of review sentence representations. All hyperparameters were tuned on the development set (see Appendix~\ref{sec:implementation} for more details).

\begin{table}[t!]
    \small
    \centering
	\scriptsize
\resizebox{0.48\textwidth}{!}{
		\begin{tabular}{ l|c|c|c} 
			\hline
			& {Reviews} & {Train / Test Ent.} & {Rev./Ent.} \TBstrut\\ 			
			\hline
			\textsc{Space} &  1.14M & 11.4K / 50 & 100\Tstrut\\
			\textsc{Amazon} & 4.75M & 183K / 60 & 8\Bstrut\\
			\hline
	\end{tabular}
}
\caption{ Dataset statistics for \textsc{Space} and \textsc{Amazon} datasets. (Train/Test Ent.: Number of entities in the \textit{training} and \textit{test} set; Rev./Ent.: Number of reviews per entity in the \textit{test} set.)}
\vspace{-8pt}
\label{tab:stat}
\end{table}

\subsection{Metrics}
\label{sec:metrics}
We report ROUGE F-scores that compares the overlap between generated text with gold summaries. 
For \textsc{Space} dataset, we measure how much general summaries cover different aspects by computing the mean ROUGE-L score with the gold aspect summaries (denoted by RL\textsubscript{ASP}). 

{We also} compute \textit{perplexity} (PPL) score to evaluate the readability of summaries. Perplexity is computed using cross-entropy loss from a BERT-\textit{base} model. We measure \textit{aspect coverage} of a system, by  computing the average number of \textit{distinct aspects} $N_{\text{ASP}}$ in the generated summaries. Lastly, to evaluate \textit{repetition} in summaries, we compute the percentage of distinct $n$-grams ($n=2$).

\subsection{Baselines}

{Following prior work~\cite{angelidis2021extractive}, we compare {\ST} with three types of systems:}
    
    \noindent(a) \textit{Best Review} systems: 
    We report the performance of \textit{Centroid} method, where reviews are encoded using BERT or SentiNeutron \cite{radford2017learning}, and the review most similar to the mean representation is selected.
    
    \noindent(b) \textit{Abstractive} systems: 
    We report the performance of \textit{Opinosis} \cite{ganesan2010opinosis} (a graph-based approach), \textit{MeanSum} \cite{chu2019meansum}, \textit{CopyCat} \cite{bravzinskas2019unsupervised} and \textit{AceSum} \cite{amplayo-etal-2021-aspect} summarization models.
    
    \noindent(c) \textit{Extractive} systems: 
    We report the performance of \textit{LexRank} \cite{erkan2004lexrank}, {where sentences were encoded using BERT, SentiNeutron or tf-idf vector.} We also report the performance achieved by selecting review sentences \textit{randomly}.

\subsection{Results}
\textbf{General Summarization}: We present the results of general summarization on \textsc{Space} dataset in Table~\ref{tab:space-results}.  {\ST} and its variants show strong improvements over previous state-of-the-art QT, and other baselines, across all ROUGE metrics.  They also outperform abstractive systems (like CopyCat and Meansum)  by a large margin, which shows that {\ST} can effectively select relevant sentences from a large pool of reviews. 
All variants of {\ST}  outperform other models in RL$_{\mathrm{ASP}}$ metric, showcasing that general summaries from {\ST} cover aspects better than baselines. {{We compiled some baseline results from \citet{angelidis2021extractive}.}}

We further evaluate the quality of the summaries, for all variations of {\ST} along with our strongest baseline QT, using other automatic metrics in Table~\ref{tab:ablations}. 
The first row in Table~\ref{tab:ablations}  reports the performance of QT, which achieves the highest distinct $n$-gram score, but has poor perplexity score. This shows that QT generates summaries with diverse text but they are not coherent. {\ST} achieves the best perplexity score (second row in Table~\ref{tab:ablations}) but produces less diverse text (lowest distinct $n$-gram score). The third row in Table~\ref{tab:ablations} reports the performance of {\ST} with redundancy term. Comparing rows 2 and 3 of Table~\ref{tab:ablations}, we observe that the summaries from {\ST} (w/ redundancy) have more distinct $n$-grams (less repetition), while falling behind in perplexity and aspect coverage.  Performance results for aspect-aware variants of {\ST} are reported in last two rows of Table~\ref{tab:ablations}. 
We observe that iteratively covering aspects  reduces repetition 
(increase in distinct-$n$ score). As expected the mean aspect-coverage ($\mathbb{E}[N_{\mathrm{ASP}}]$) improves in aspect-aware {\ST} variants. However, a slight drop in aspect-coverage is observed when the redundancy term is introduced (last row in Table~\ref{tab:ablations}). We also observe an increase in perplexity for aspect-aware variants, which can be caused due to multiple changes in aspect context. Overall, {\ST} (w/ aspect + redundancy) is able to produce diverse text with a high aspect coverage and a decent perplexity score, appearing to be the best performing model. 

\begin{table}[t!]
	\centering
		\resizebox{0.5\textwidth}{!}{
	\begin{tabular}{@{}cl@{}|c c c|c} 
		\toprule[1pt]
		\multicolumn{2}{l@{~}|}{\textsc{Space} [General]} & {R1} & {R2} & {RL} & RL$_{\text{ASP}}$ \TBstrut\\ 			
		\midrule[1pt]
		\parbox[t]{0.6mm}{\multirow{4}{*}{\rotatebox[origin=c]{90}{\small{Best Review}}}} 
		& {Centroid}\textsubscript{SENTI} & 27.36 & 5.81 & 15.15 & 8.77 \Tstrut\\
		& Centroid\textsubscript{BERT} & 31.33 & 5.78 & 16.54 & 9.35\\
		& Oracle\textsubscript{SENTI} & \light{32.14} & \light{7.52} & \light{17.43} & \light{9.29}\\
		& Oracle\textsubscript{BERT} &  \light{33.21} & \light{8.33} & \light{18.02} & \light{9.67}\Bstrut\\
		\midrule[1pt]
		\parbox[t]{0.6mm}{\multirow{4}{*}{\rotatebox[origin=c]{90}{\small{Abstract}}}} 
		& Opinosis (\citeauthor{ganesan2010opinosis}) & 28.76 & 4.57 & 15.96 & 11.68\Tstrut\\
		& MeanSum (\citeauthor{chu2019meansum}) & 34.95 & 7.49 & 19.92 & 14.52\\
		& {Copycat (\citeauthor{bravzinskas2019unsupervised}) } & 36.66 & 8.87 & 20.90 & 14.15\\ 
		& {AceSum (\citeauthor{amplayo2021unsupervised}) } & 40.37 & 11.51 & 23.23 & -\\
		\midrule[1pt]
		\parbox[t]{0.6mm}{\multirow{6}{*}{\rotatebox[origin=c]{90}{\small{Extract}}}} 
		& Random & 26.24 & 3.58 & 14.72 & 11.53\Tstrut\\
		& LexRank\textsubscript{TF-IDF} & 29.85 & 5.87 & 17.56 & 11.84\\
		& LexRank\textsubscript{SENTI} & 30.56 & 4.75 & 17.19 & 12.11\\
		& LexRank\textsubscript{BERT} & 31.41 & 5.05 & 18.12 & 13.29\\& 
		{AceSum\textsubscript{EXT} (\citeauthor{amplayo2021unsupervised}) } & 35.50 & 7.82 & 20.09 & -\\
		& QT (\citeauthor{angelidis2021extractive}) & 38.66 & 10.22 & 21.90 & 14.26\Bstrut\\
    \cmidrule{2-6}
		& {\ST} &  {42.48}  & \textbf{13.48} & \textbf{26.40} & {15.23}\Tstrut\\
		& \hspace*{0.55cm}w/ redun. &  42.06 & 12.69 & 25.77 & \textbf{15.40}\\
		& \hspace*{0.55cm}w/ aspect &  42.86 & 12.92 & 25.52 & 15.22\\
		& \hspace*{0.55cm}w/ aspect + redun.\hspace*{0.08cm} &  \textbf{43.46} & 13.06 & 25.43 & 15.14\Bstrut\\
		\bottomrule[1pt]
	\end{tabular}
	}
	\vspace{-5pt}
	\caption{ Evaluation results on \textsc{Space} dataset. Best results for each metric are shown in \textbf{bold}. RL$_{\text{ASP}}$ is the average ROUGE-L score when compared with gold aspect-specific summaries. Systems that access reference summaries are reported in \light{gray}.}
	\label{tab:space-results}
	\vspace{-5pt}
\end{table}

\begin{table}[t!]
    \small
	\centering
\scriptsize
\resizebox{0.48\textwidth}{!}{
	\begin{tabular}{ l| c  c c} 
		\hline
		\textsc{Space} [General]  & {PPL} & $\mathbb{E}[N_{\mathrm{ASP}}]$ & Distinct-$n$ \TBstrut\\ 			
		\hline
		{QT}  & 4.96 & 4.40 & \textbf{0.98}\Tstrut\\
		{\ST} &  \textbf{3.37} & 4.44 & 0.89\Bstrut\\
		\hspace*{0.15cm}w/ redun.   & 4.01 & 4.12 & 0.93\\
		\hspace*{0.15cm}w/ aspect & 3.55 & \textbf{5.24} & 0.94\\ 
		\hspace*{0.15cm}w/ aspect + redun. & 3.70 & 4.84  & 0.95 \\ 
		\hline
	\end{tabular}
	}
    \vspace{-5pt}
	\caption{ Evaluation results of QT,  {\ST} and its different variations on \textsc{Space} general summarization. For all setups with redundancy term constant $\gamma=0.1$.}
    \vspace{-15pt}
	\label{tab:ablations}
\end{table}

Evaluation results on \textsc{Amazon} dataset are reported in Table~\ref{tab:amazon-results}. {\ST} and its variants\footnote{We do not have aspect-aware selection variants in \textsc{Amazon}, as it does not provide aspect-denoting keywords.} achieve similar performance, with {\ST} {achieving the best performance among all extractive summarization system}.
{{\ST} falls short of only abstractive summarization systems that have the advantage of generating novel phrases not present in the input reviews. }
Also, while {\ST} {beats most baselines} for \textsc{Amazon} dataset, the performance gain isn't as much as \textsc{Space} dataset. We believe this is because the number of reviews per entity in \textsc{Amazon} (8) is much lower compared to \textsc{Space} (100).    
As {\ST} is dependent on the mean representation $\bar{\alpha}$, having more reviews helps in capturing the popular opinion distribution accurately.\footnote{{We observed a drop in performance when the number of reviews/entity in \textsc{Space} dataset was reduced (experimental details in Section~\ref{sec:num-reviews}).} \\ $^{\ssymbol{2}}$
\footnotesize{Reported results are obtained using the publicly released implementation of QT \cite{angelidis2021extractive}.}} 
For practical purposes, opinion summarization systems are useful when there are hundreds or more reviews per entity. A larger improvement on \textsc{Space} shows the efficacy of {\ST} in the real world. 

\begin{table}[t!]	
    \centering
	\resizebox{0.4\textwidth}{!}{
	\begin{tabular}{@{}cl@{}|c c c} 
		\toprule[1pt]
		\multicolumn{2}{l@{~}|}{\textsc{Amazon}} & {R1} & {R2} & {RL} \TBstrut\\ 			
		\midrule[1pt]
		\parbox[t]{1mm}{\multirow{3}{*}{\rotatebox[origin=c]{90}{\small{Best Rev.}}}} 
		& Random & 27.66 & 4.72 & 16.95\Tstrut\\
		& 		Centroid\textsubscript{BERT}\hspace*{0.08cm} & 29.94 & 5.19 & 17.70 \\
				& Oracle\textsubscript{BERT} & \light{31.69} & \light{6.47} & \light{19.25} \Bstrut\\
		\midrule[1pt]
		\parbox[t]{1mm}{\multirow{6}{*}{\rotatebox[origin=c]{90}{\small{Abstract}}}} 
		& {Opinosis (\citeauthor{ganesan2010opinosis})} & 28.42 & 4.57 & 15.50 \Tstrut\\
		& {MeanSum (\citeauthor{chu2019meansum})} & 29.20 & 4.70 & 18.15 \\
		& {CopyCat (\citeauthor{bravzinskas2019unsupervised}) }  & {31.97} & {5.81} & {20.16} \\
		& {PlanSum (\citeauthor{amplayo2021unsupervised}) } & 32.87 & 6.12 &  19.05\\
		& {TranSum (\citeauthor{wang-wan-2021-transsum}) } & 34.23 & \underline{7.24} & 20.49\\
		& {\textsc{Coop} (\citeauthor{iso-etal-2021-convex-aggregation}) } &  \underline{36.57} & {7.23} & \underline{21.24} \Bstrut\\
		\midrule[1pt]
		\parbox[t]{1mm}{\multirow{3}{*}{\rotatebox[origin=c]{90}{\small{Extract}}}}
		& {LexRank\textsubscript{TF-IDF}} \hspace*{0.08cm} & 28.56 & 3.98 & 15.29 \Tstrut\\
		& LexRank\textsubscript{BERT} & 31.47 & 5.07 & {16.81}\\
		& 		QT$\ssymbol{2}$ (\citeauthor{angelidis2021extractive}) & {31.27} & 5.03 & 16.42 \Bstrut\Bstrut\\
		\cline{2-5}
				& {\ST} & \textbf{32.03} & 5.38 & 16.47\Tstrut\\
				& \hspace*{0.55cm}w/ redun.\hspace*{0.15cm} & {31.92} & \textbf{5.68} & \textbf{16.61}\Bstrut\\
		\bottomrule[1pt]
	\end{tabular}
	}
	\caption{ Evaluation results on \textsc{Amazon} dataset. Best performance achieved using an extractive systems are in \textbf{bold}. Overall best results  for  each  metric  is \underline{underlined}.  System performance that access reference summaries are reported in \light{gray}.}
	\label{tab:amazon-results}
	\vspace{-15pt}
\end{table}

\begin{table*}[t!]	
\centering
	\resizebox{\textwidth}{!}{
	\begin{tabular}{ l|c c c c c c | c c c} 
		\toprule[1pt]
		\textsc{Space} [Aspect] & Building & Cleanliness & Food & Location & Rooms & Service & $\overline{\text{R1}}$ & $\overline{\text{R2}}$ & $\overline{\text{RL}}$\TBstrut\\ 			
		\hline
		MeanSum (\citeauthor{chu2019meansum})
		& 13.25 & 19.24 & 13.01 & 18.41 & 17.81 & 20.40 & 23.24 & 3.72 & 17.02\Tstrut\\
		CopyCat (\citeauthor{bravzinskas2019unsupervised})
		& 17.10 & 15.90 & 14.53 & 20.31 & 17.30 & 20.05 & 24.95 & 4.82 & 17.53\\
		LexRank\textsubscript{BERT} (\citeauthor{erkan2004lexrank}) 
		& 14.73 & 25.10 & 17.56 & 23.28 & 18.24 & 26.01 & 27.72 & 7.54 & 20.82 \\
		QT (\citeauthor{angelidis2021extractive})
		& 16.45 & \textbf{25.12} & 17.79 & {23.63} & 21.61 & {26.07} & 28.95 & 8.34 & 21.77\Bstrut\\
		\hline
		{\ST} 
		& \textbf{20.04} & 23.72 & \textbf{23.57} & \textbf{25.33} & \textbf{25.29} & \textbf{26.90} & \textbf{31.24} & \textbf{10.43} & \textbf{24.14}\Tstrut\\
		\hspace*{0.2cm}w/o  informativeness & 18.38 & 24.08 & 19.03 & 23.32 & 23.89 & 25.05 & 27.85 & 8.61 & 22.29 \Bstrut\\
			\bottomrule[1pt]
	\end{tabular}
}
	\vspace{-5pt}
	\caption{ Evaluation results of Aspect Summarization on \textsc{Space} dataset. ROUGE-L scores are reported for six different aspects. $\overline{\text{R1}}$, $\overline{\text{R2}}$ and $\overline{\text{RL}}$ are the average ROUGE-1, ROUGE-2 and ROUGE-L F scores respectively. Best system results are in \textbf{bold}. 
	}
	\label{tab:aspect-results}
	\vspace{-10pt}
\end{table*}

\noindent\textbf{Aspect Summarization}: For aspect summarization, we compare against four unsupervised systems MeanSum, CopyCat, LexRank and QT on the \textsc{Space} dataset. 
For general summarizers: MeanSum, CopyCat and LexRank, sentence embeddings retrieved from BERT \cite{vaswani2017attention} were clustered using $k$-means and each cluster $S_e^{(a)}$ was assigned an aspect $a$ based on frequency of {aspect-denoting} keywords 
in the cluster's sentences. The models  then produced summaries for each aspect $a$ given the input set $S_e^{(a)}$. All models 
 including {\ST}, use the same {aspect-denoting} keywords. 

Evaluation results on \textsc{Space} are reported in Table~\ref{tab:aspect-results}. {\ST} outperforms the state-of-the-art QT in all aspects except \textit{cleanliness}, where the performance is comparable. We observe that adding the \textit{informativeness} term ($\Delta(\bar{\alpha},  \alpha^s)$ in  Equation~\ref{eqn:aspect-inf}) 
 helps improve the specificity of the aspect thereby boosting performance. {\ST} also shows significant gains in terms of average ROUGE-1/2 and ROUGE-L across different aspects.

\begin{table}[t!]
	\centering
    \resizebox{0.48\textwidth}{!}{
	\begin{tabular}{ l| c  c c} 
		\toprule[1pt]
		\textsc{Space} [General]  & {Inform.} & {Coherence} & {Redund.} \Tstrut\\ 			
		\midrule[1pt]
        {QT}  & -31.3 & -47.3 & -39.3\\
		{\ST} (w/ asp. + redun.) & \multirow{1}{*}{\textbf{-21.3}*} & \multirow{1}{*}{\textbf{-28.0}*} & \multirow{1}{*}{\textbf{-27.3}*} \\
		\light{Human} & \light{+52.7} & \light{+75.3} & \light{+66.7} \Bstrut\\
		\bottomrule[0.5pt]
	\end{tabular}
}

    \resizebox{0.48\textwidth}{!}{
	\begin{tabular}{l| c  c } 
		\hline
		\textsc{Space} [Aspect]  & {Asp. Inform.} & {Asp. Specificity} \Tstrut\\ 			
		\midrule[1pt]
        {QT} & -35.0 & -24.7 \\		
		{\ST}\hphantom{ (w/ asp. + redun.)}	& \textbf{-13.0}* & \textbf{-11.0} \\
		\light{Human} & \light{+48.0} & \light{+35.7} \Bstrut\\
		\bottomrule[1pt]
	\end{tabular}
	}
	\caption{Human evaluation results of general and aspect summarization for \textsc{Space} dataset. Best human evaluation results obtained for a system  are in \textbf{bold} and human performance is in \light{gray}. (*): statistically significant difference with QT model ($p < 0.05$, using paired bootstrap resampling \citet{koehn2004statistical}).}
	\label{tab:human-eval-general}
	\vspace{-15pt}
\end{table}

\noindent\textbf{Human Evaluation}: 
We performed human evaluations for the general and aspect summaries.  We evaluated general summaries from QT, best performing variant {\ST} (w/ aspect + redundancy) and gold summary. Summaries were judged by 3 human annotators on three criteria: \textit{informativeness}, \textit{coherence} and \textit{non-redundancy}.  The judges were presented summaries in a pairwise manner and asked to select which one was better/worse/similar. The scores (-100 to +100) were computed using \textit{Best-Worst Scaling} \cite{louviere2015best}. The first half of Table~\ref{tab:human-eval-general} reports the evaluation results, where we observe that {\ST} (w/ aspect + redundancy) outperforms our strongest baseline, QT, for all criteria (statistical significance information provided in the caption of Table~\ref{tab:human-eval-general}). However, summaries generated from both systems are far from gold summaries on all criteria.

We also evaluated aspect summaries generated by {\ST} and QT in a similar manner. Aspect summaries were judged based on two criteria: \textit{aspect informativeness} (usefulness of opinions for a specific aspect, consistent with reference) and \textit{aspect specificity} (how specific the summary is for an aspect without considering other factors). The bottom half of Table~\ref{tab:human-eval-general} reports the results for aspect summaries. We observe that both QT and {\ST} produce aspect-specific summaries. However, {\ST} shows a statistically significant improvement over QT in aspect informativeness.

\subsection{Analysis}
\label{sec:analysis}

\begin{figure}[t!]
	\centering
	\includegraphics[width=0.25\textwidth]{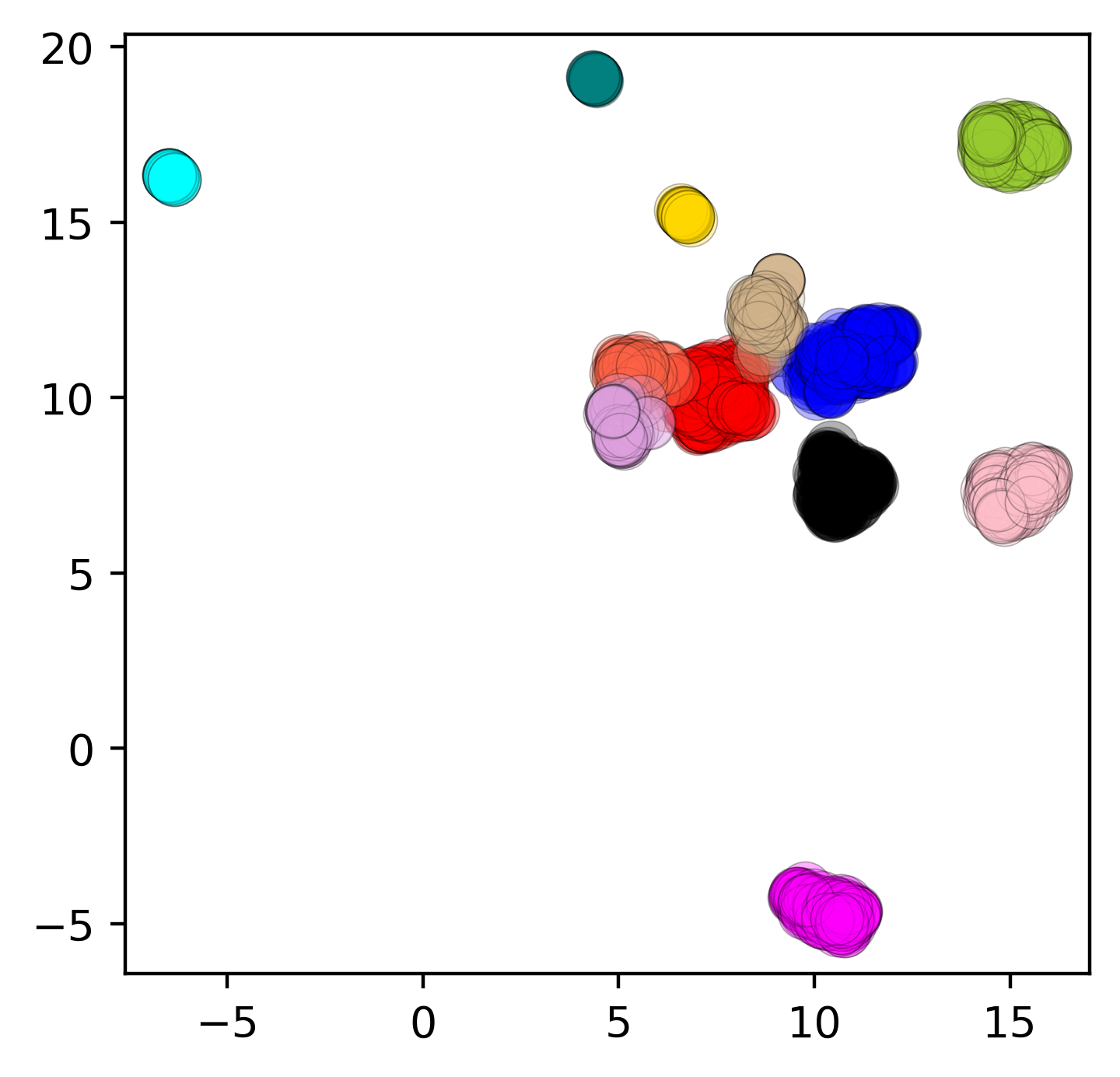}
	\vspace{-5pt}
	\caption{ Visualization of UMAP  projections of dictionary elements. Projections form clusters, which are shown in different colors. } 
	\label{fig:dict-viz}
	\vspace{-15pt}
\end{figure}

\noindent\textbf{Latent Dictionary Interpretation.} In this section, we investigate the semantic meanings learnt by individual dictionary elements, $D_k$. 
We visualized the UMAP projection \cite{mcinnes2018umap} of dictionary element representations (shown in Figure~\ref{fig:dict-viz}). For different runs of {\ST}, we found that the dictionary representations converged into clusters as shown in Figure~\ref{fig:dict-viz} (elements are color-coded according to their cluster identities as assigned by $k$-means algorithm with $k$=12). 

\begin{table*}[t!]
    \footnotesize
    \centering

\resizebox{0.9\textwidth}{!}{
\begin{tabular}{p{0.233\textwidth} | p{0.225\textwidth} | p{0.225\textwidth} | p{0.235\textwidth}} 
		\toprule[1pt]
		\multicolumn{1}{c|}{{\bf \ST}} & \multicolumn{1}{c|}{{\bf \ST} (w/ redun.)} & \multicolumn{1}{c|}{{\bf \ST} (w/ aspect)} & \multicolumn{1}{c}{{\bf \ST} (w/ aspect + redun.)} \TBstrut\\ 			
		\midrule[1pt]
		& & & \\[-0.9em]
		{\sethlcolor{LightRed}\hl{The {\bf staff} is great.}}	The Hotel Erwin is a great {\bf place} to stay.	{\sethlcolor{LightRed}\hl{ The {\bf staff} were friendly and helpful.	The location is perfect}}.	We ate {\bf breakfast} at the hotel and it was great.	{\sethlcolor{LightRed}\hl{The hotel itself is in a great \textbf{location}. The \textbf{service} was wonderful.}}	It was great.	The {\bf rooms} are great.	The rooftop {\bf bar} HIGH was the icing on the cake.	The {\bf food} and {\bf service} at the restaurant was awesome.	{\sethlcolor{LightRed}\hl{The \textbf{service} was excellent.}}
		
		& {The hotel itself is in a great  {\bf \bf location}.}	The  {\bf rooms} were clean and we were on the 5th.	The best part of the  {\bf hotel} is the 7th floor rooftop deck.	The  {\bf staff} is great.	The  {\bf hotel} has so many advantages over the other options in the area that it is a no contest.	If you want to stay in Venice, this is a great  {\bf place} to be.	The  {\bf food} and  {\bf service} at the restaurant was awesome.
		
		&  	{\sethlcolor{LightRed}\hl{The  {\bf staff} is great.	The  {\bf staff} were friendly and helpful.}}	The Hotel Erwin is a great  {\bf place} to stay.	The  {\bf location} is perfect.	We ate  {\bf breakfast} at the hotel and it was great.	The  {\bf food} and  {\bf service} at the restaurant was awesome.	{\sethlcolor{LightRed}\hl{The  {\bf rooms} are great.	The  {\bf room} is epic!}}	The rooftop  {\bf bar} HIGH was the icing on the cake.	The rooftop  {\bf bar} at the hotel, "High", is amazing. 
		
		& {The  {\bf staff} is great.	We had a great stay at the Erwin, and the  {\bf staff} really made it more enjoyable.	The Hotel Erwin is a great  {\bf place} to stay.	It was great.	We ate  {\bf breakfast} at the hotel and it was great.	The  {\bf food} and  {\bf service} at the restaurant was awesome.	The  {\bf rooms} are great.	We had a  {\bf kitchen and balcony} and partial ocean view.	The rooftop  {\bf bar} HIGH was the icing on the cake.}
		\Tstrut\\
		\bottomrule[1pt]
\end{tabular}
}
	\vspace{-5pt}
	\caption{ Example summaries from different variants of {\ST}. Redundant sentences are {\sethlcolor{LightRed}\hl{highlighted}}. The aspect denoting words are in \textbf{bold}. For {\ST} \& {\ST} (w/ redun.), we observe frequent context switch among aspects. {\ST} (w/ aspect) \& {\ST} (w/ aspect + redun.) summaries cover different aspects in a coherent manner.}
	\label{tab:ablated-summaries}
	\vspace{-10pt}
\end{table*}

\begin{table}[t!]
	\footnotesize
	\centering
	\resizebox{0.48\textwidth}{!}{
\begin{tabular}{p{0.05\textwidth} | p{0.27\textwidth} | p{0.09\textwidth}  } 
		\toprule[1pt]
		 $(h, k)$ & \multicolumn{1}{c|}{Sentences w/ high activation} & \multicolumn{1}{c}{Explanation}\TBstrut\\ 			
		\midrule[1pt]
		& & \\[-0.75em]
		$(3, 5)$ 
		& 
			\tabitem I wish all hotels or any business for that matter, had employees a dedicated to service as he was.
			
			\tabitem Very polite and very professional approach.
		& Service \TBstrut\\
		\hline & & \\[-0.75em]
		$(0, 10)$
		 & \tabitem Stayed here in August for the our first trip to Vancouver.
		
		\tabitem I stayed at this motel with my partner in August 2010. 
		& Phrase ``stayed''\TBstrut\\
		\hline & & \\[-0.75em]
		 
		$(6, 0)$ & 
		\tabitem Empty water bottles were never thrown out and no one put the iron and ironing board away.
		
		\tabitem Facing St Paul St can be a very noisy experience.
		& Bad experience\TBstrut\\
		\hline & & \\[-0.75em]
		(2, 8) & 
		\tabitem A full cooked to order breakfast (including omlettes,  \ldots, fruit, etc.)
		
		\tabitem Pizza hut, Mc donalds, KFC all round the corners...
		& Food \TBstrut\\
		\hline & & \\[-0.75em]
		(5, 8) &
	    \tabitem The rooms seem small, tight fit for a family of 4. 
	    
	    \tabitem You may have a difficult fit.
	    & Small rooms\TBstrut\\
		\bottomrule[1pt]
\end{tabular}
}
	\vspace{-5pt}
	\caption{List of sentences with high activation value with cluster means of dictionary elements. For each head representation, cluster means capture different semantics. $h$: head index; $k$: cluster index.} 
	\vspace{-15pt}
	\label{tab:patterns}
\end{table}

We hypothesize that the clusters 
should capture certain semantic meaning. We explore this hypothesis by 
identifying sentences sharing similar representations with the mean representations $\{\mu_1, \ldots, \mu_K\}$ for each cluster. 
For each head $h$ in the encoder (Section~\ref{sec:encoder}), we compute cosine similarity of sentences with cluster means. Table~\ref{tab:patterns} shows some examples of sentences having highest similarity with a cluster mean $\mu_k$ for a head representation $h$. 
We observe in most cases sentences closest to a cluster share a similar semantic meaning. For hotel reviews, we observe that sentences often talk about a specific aspect like service, food and rooms, as shown for $(h, k)$ configurations (3, 5), (2, 8) and (5, 8) in Table~\ref{tab:patterns}. 
The clusters sometimes capture certain coarse semantics like presence of a word or  phrase (e.g. config. (0, 10) in  Table~\ref{tab:patterns}). It can also capture high-level semantics like the experience of a customer (e.g. config.  (6, 0)).  It was interesting to observe that a single cluster can capture different semantics for distinct heads (cluster 8 in configurations (2, 8) and (5, 8)). 

\noindent\textbf{Qualitative Examples.} Table~\ref{tab:ablated-summaries} shows 
summaries 
generated by {\ST} and its variants for the \textsc{Space} dataset. 
	 While {the} summary generated by {\ST}
	talks about {\textit{location}, \textit{staff} \& \textit{service} multiple times} (shown as {\sethlcolor{LightRed}\hl{highlighted}} text), summary from {\ST} (w/ redundancy) doesn't have that repetition.   

Also, the summary generated by {\ST}  switches context frequently. For example, the aspect of the first three sentences changes from service$\rightarrow$location$\rightarrow$service.
We observe that compared to {\ST}, both aspect-aware 
{\ST} variants generate summaries without abrupt context switches. 
The summary generated by {\ST} (w/ aspect) covers aspects like service, hotel, food and rooms sequentially, but sentences referring to an aspect are quite similar. {\ST} (w/ aspect + redundancy) overcomes this shortcoming, and introduces diversity among the aspect-specific sentences.

\begin{table}[t!]
	\centering
	\resizebox{0.4\textwidth}{!}{
	\begin{tabular}{ l|c c c c} 
		\toprule[1pt]
		\textsc{Space} [General] & {5\%} & {10\%} & {50\%} & 100\% \TBstrut\\ 			
		\midrule[1pt]
	Copycat & 26.1 & 26.2 & 31.8 & 36.7 \Tstrut\\
		QT & 36.9 & 37.1 & 37.7 & 38.7 \Bstrut\\
		\midrule[1pt]
		{\ST} & 37.8 & 40.9 & 41.2 & 42.5\TBstrut\\
		\bottomrule[1pt]
	\end{tabular}
}
	\caption{ ROUGE-1 scores with different training data.}
	\vspace{-15pt}
	\label{tab:train-size}
\end{table}

\noindent\textbf{Training Data Efficiency.} 
We analyze the performance of {\ST}, QT and CopyCat for general summarization (ROUGE-1) on \textsc{Space} for varying training data fractions  in Table~\ref{tab:train-size}. We observe that both QT and {\ST} perform well with low training data. 
However, {\ST} outperforms QT in all low resource settings. {\ST} (with 10\% data) yields significant ROUGE-1 improvements over QT (with access to 100\% data).

\begin{table}[t!]
	\centering
	\resizebox{0.3\textwidth}{!}{
	\begin{tabular}{ c |c c c} 
		\toprule[1pt]
		Rev./Ent. & {R1} & {R2} & {RL} \TBstrut\\ 			
		\midrule[1pt]
    	5 & {40.49} & 12.92 & 26.23 \Tstrut\\
		10 & 40.76 & 13.14 & 26.26  \\
		25 & 41.17 & 13.18 & 26.05  \\
		50 & 41.55 & 13.16 & 26.01\\
		\hline
		100 & \textbf{42.48} & \textbf{13.48} &  \textbf{26.40} \TBstrut\\
		\bottomrule[1pt]
	\end{tabular}
}
	\vspace{-5pt}
	\caption{ ROUGE-F scores of {\ST} with varying number of reviews per entity. }
	\label{tab:num-reviews}
	\vspace{-15pt}
\end{table}

\noindent{\textbf{Impact of number of reviews.}\label{sec:num-reviews} 
We investigate whether {\ST}'s performance gain on {\textsc{Space}}  is due to the larger number of reviews available (reviews per entity  
-- \textsc{Amazon}: 8, \textsc{Space}: 100). Specifically, we perform ablation experiments by reducing the number of reviews/entity in \textsc{Space} dataset. We remove user reviews with low relevance scores (relevance score of a review is the average $\mathcal{R}(\cdot)$ of its sentences). Table~\ref{tab:num-reviews} reports the performance of {\ST} with different number of reviews/entity in the test set. We observe a gradual decline in ROUGE-1 score when the reviews/entity is reduced, which shows that having more reviews per entity helps in better extractive summarization.
}

\noindent{\textbf{Additional Controllable Summarization}.\label{sec:control-summ} We showcase that {\ST} can perform different forms of controllable summarization. Specifically, we perform sentiment-based summarization using a small number (10) of seed sentences belonging to \textit{positive}, \textit{negative} and \textit{neutral} sentiment class. Seed sentences were annotated using the rule-based system VADER \cite{hutto2014vader}. An example of sentiment-based summarization is shown in Table~\ref{tab:custom-summ}. We observe {\ST} is able to generate summaries aligning with the seed sentiments.} We also perform multi-aspect summarization using {\ST}, by controlling the aspect of the selected sentences. Table~\ref{tab:multi-aspect} showcases an example of multi-aspect summarization. An interesting observation is that SemAE is able to select sentences, which  have mutliple aspects (shown in \textcolor{blue}{\bf blue}) and not independent sentences from different aspects. These experiments show that SemAE is able capture and leverage granular semantics for summarization.

\begin{table}[t!]
    \footnotesize
	\centering
	\resizebox{0.47\textwidth}{!}{
	\begin{tabular}{c| p{0.35\textwidth}}
		\toprule[1pt]
		\textsc{Sentiment} & \multicolumn{1}{c}{\textsc{Summary}} \TBstrut\\ 			
		\midrule[1pt]
	 	\textcolor{black}{{Positive}} & \textcolor{blue}{\bf {Love} the warm chocolate chips cookies and the service has always been outstanding.} Excellent morning breakfasts and the airport shuttle runs every 15 minutes but we have made the 10 minute walk numerous times to the airport terminal. \TBstrut\\
		\midrule[0.5pt]\\[-1em]
		\textcolor{black}{{Negative}} & To add insult to injury, for people who use the parking lot to "park and fly", the charge is \$7.95/day, almost half of what the hotel guests are charged!! \textcolor{red}{\bf \textbf{Cons} - Hotel is spread out so pay attention to how to get to your room as you may get lost}, Feather pillows (synthetic available on request), Pay parking (\$16 self/day \$20 valet/day), warm cookies on check in. \TBstrut\\
		\midrule[0.5pt]\\[-1em]
		\textcolor{black}{{Neutral}} & Stayed at this hotel beause the park n fly. \textcolor{teal}{\bf We have stayed at this hotel several times in the family suite ( 2 bedrooms/1 king and 2 queen beds).} Despite the enormity of this hotel, it very much feels almost family run.\TBstrut\\
		\bottomrule[1pt]
	\end{tabular}
}
	\vspace{-5pt}
	\caption{An example of sentiment-based summarization for a hotel entity in \textsc{Space} dataset. }
	\vspace{-10pt}
	\label{tab:custom-summ}
\end{table}

\begin{table}[t!]
    \footnotesize
	\centering
	\resizebox{0.47\textwidth}{!}{
	\begin{tabular}{p{0.1\textwidth}| p{0.33\textwidth}}
		\toprule[1pt]
		\textsc{Aspects} & \multicolumn{1}{c}{\textsc{Summary}} \TBstrut\\ 			
		\midrule[1pt]
	 	({food}, {staff}) & {\textcolor{blue}{\bf The \textbf{staff} was friendly and helpful and we enjoyed the warm, chocolate chip \textbf{cookie} we were given at check-in.}}  The \textbf{breakfast} in the restaurant was amazing, and the \textbf{staff} was very attentive. 
	 	\TBstrut\\
	 	\hline\\[-0.5em]
	 	({room}, {cleanliness}) & {\textcolor{blue}{\bf The bed was very nice, \textbf{room} was \textbf{clean}, we even had a balcony.}} The beds were comfortable and the room was very \textbf{clean}.\TBstrut\\
		\bottomrule[1pt]
	\end{tabular}
}
	\vspace{-5pt}
	\caption{Examples of multi-aspect summarization for a hotel entity in \textsc{Space} dataset.}
	\vspace{-15pt}
	\label{tab:multi-aspect}
\end{table}

In Appendix~\ref{sec:extended-analysis}, we perform additional analysis to investigate the head-wise analysis, efficacy of sparsity constraints, dictionary evolution, and qualitatively compare 
{\ST} with baselines (QT and CopyCat).

\section{Conclusion}
We proposed a novel opinion summarization approach using {\SemAE}, which encodes text as a representation over latent semantic units. We perform extractive summarization by selecting sentences using information-theoretic measures over representations obtained from {\ST}. 
Our experiments reveal that dictionary element representations from {\ST} form clusters, which capture distinct 
semantics. Our model provides fine-grained control to users to model surface-level text attributes (like redundancy, informativeness etc.) in the representation space. {\ST} outperforms existing extractive opinion summarization methods on \textsc{Space} and \textsc{Amazon} datasets. Finally, {\ST} representations can be leveraged to explore different forms of control on the summary generation (e.g. multi-aspect sumamrization) using our inference framework.
Future works can focus on better representation learning systems to handle use-cases with noisy or sparse textual data. 

\section*{Acknowledgements}
{This work was supported in part by NSF grants IIS2112635 and IIS2047232. We also thank the anonymous reviewers for their thoughtful and constructive comments.}

\section*{Ethical Considerations}
We do not foresee any ethical concerns from the technology presented in this work. We used publicly available datasets, and do not annotate any data manually. The datasets used have reviews in English language. Human evaluations for summarization were performed on Amazon Mechanical Turks (AMT) platform. Human judges were compensated at a wage rate of \$15 per hour.

\bibliography{anthology,custom}
\bibliographystyle{acl_natbib}

\appendix
\clearpage
\section{Appendix}
\label{sec:appendix}


\subsection{Implementation Details}
\label{sec:implementation}
The Transformer is trained without the dictionary learning reconstruction for 4 warmup epochs.  We tokenized text in an unsupervised manner using SentencePiece\footnote{https://github.com/google/sentencepiece} tokenizer with 32K vocabulary size. The model was trained using Adam Optimizer with a learning rate of $10^{-3}$, and a weight decay of 0.9. Our model was trained for 10 epochs on a single GeForce GTX 2080 Ti GPU in 35 hours. The loss function parameters are reported in Table~\ref{tab:hyperparams}. { The hyperparameters were tuned on the development set of the dataset based on ROUGE-1 F score. For aspect summarization, we set $\beta=0.7$ after tuning (grid search between 0.1 and 1, with intervals of 0.1) on the development set.} We choose the redundancy term constant $\gamma = 0.1$ in a similar manner. Post training, the summaries were generated with $N=20$. We limit the summary length to 75 tokens. Each keyword $w_i \in Q_a$ is associated with a confidence score for aspect $a$. In case a sentence has multiple keywords belonging to different aspects we use the confidence score to assign the aspect.

\begin{table}[t!]
	\centering
	\resizebox{0.25\textwidth}{!}{
	\begin{tabular}{ l|c c} 
		\toprule[1pt]
		\textsc{Dataset} & {$\lambda_1$} & {$\lambda_2$} \TBstrut\\ 			
		\midrule[1pt]
		\textsc{Amazon} & $10^3$ & $5 \times 10^{-4}$\\
		\textsc{Space} & $10^4$ & $5 \times 10^{-4}$ \\
		\bottomrule[1pt]
	\end{tabular}
}
	\caption{Loss function hyperparameters values.}
	\label{tab:hyperparams}
\end{table}

\subsection{Dataset Construction}
\label{sec:dataset}

{
In this section, we provide some background information about the dataset creation process for \textsc{Space} and \textsc{Amazon}. \textsc{Space} corpus has a large number of reviews per entity. Therefore, \citet{angelidis2021extractive} collected summaries from reviews following a two-step procedure (a) sentence voting, and (b) summary collection. \textit{Sentence voting} step involves selecting informative review sentences using a majority vote from the annotators. Annotators were prompted to select between 20-40\% of the total sentences. \textit{Summary collection} involves generating a overview summary of the selected sentences upto a 100-word budget. For aspect summaries, selected sentences were annotated using an off-the-shelf aspect classifier \cite{angelidis2018summarizing}. Human annotators were asked to summarize selected sentences belonging to an aspect. \textsc{Amazon} dataset has a relatively lower number of reviews per entity. The evaluation set of \textsc{Amazon} was created by sampling 60 entities and 8 reviews per entity. These were provided to the human annotators for summarization \cite{bravzinskas2019unsupervised}.}

\subsection{Ablations}
\label{sec:ext-analysis}
\begin{itemize}[leftmargin=*, topsep=0pt]
\begin{table}[t!]
	\centering
    \resizebox{0.5\textwidth}{!}{
	\begin{tabular}{ l|c c c| c c c} 
		\toprule[1pt]
		\textsc{Space} & {R1} & {R2} & {RL} & PPL & $\mathbb{E}(N_{ASP})$ & Dist. $n$ \TBstrut\\ 			
		\midrule[1pt]
		{\ST} &  {42.48}  & {13.48} & {26.40} & {3.37} & {4.44} & {0.89} \TBstrut\\
		\hline
		\hspace*{0.08cm}w/ cosine $\Delta$ & 42.53 & 13.67 & 26.12 & 3.41 & 4.44 & 0.89\Tstrut\\
		\hspace*{0.08cm}w/ inform.  & 42.48 & 13.47 & {26.13} & 3.32 & 4.44 & 0.89\\ 
		\bottomrule[1pt]
	\end{tabular}
}
	\caption{Evaluation results of ablation experiments. For informativeness term, $\beta' = 0.1$.}
	\label{tab:ext-ablations}
\end{table}
	\item   \textbf{Divergence metric}:
	{\ST} uses KL divergence to measure the relevance of a sentence $\alpha^s$ when compared to the mean $\bar{\alpha}$, we used KL-divergence earlier. In this setup, we experiment with cosine similarity as our divergence function $\Delta(\cdot, \cdot)$. 
	The modified divergence $\Delta(\cdot, \cdot)$ score is defined as:
	\begin{equation}
		\Delta(\alpha^s, \bar{\alpha}) = \sum_h \frac{\bar{\alpha}_h^{T} \alpha_h^s}{\lvert\lvert \bar{\alpha}_h\rvert\rvert_2 \lvert\lvert \alpha_h^s \rvert\rvert_2}
	\end{equation}
	The second row in Table~\ref{tab:ext-ablations} reports the performance in this setup, which is similar to the baseline {\ST} performance. This shows that cosine similarity can serve as a good proxy to measure relevance $\mathcal{R}(\cdot)$.

	\item  \textbf{Informativeness}:\label{sec:inf} In this ablation experiment, we incorporate the informativeness term 
	in general summarization. The modified relevance score is: 
	\begin{equation}
	\mathcal{R}(\alpha^s) = \Delta(\bar{\alpha}, \alpha^s) - \beta'\Delta(\alpha^{(b)}, \alpha^s)
	\label{eqn:gen-inf}
	\end{equation}
	where $\alpha^{(b)} = \mathop{\mathbb{E}}[\alpha^s]$, the mean representation of all sentences across all entities. $\alpha^{(b)}$ captures \textit{background knowledge} distribution \cite{peyrard2018simple}, and a good summary should be divergent from the background information. Third row in Table~\ref{tab:ext-ablations} reports the performance in this setup, where we do not observe any gain over the baseline. We believe this maybe due to the fact that $\alpha^{(b)}$ doesn't capture the background knowledge properly, as it is the mean representation of hotel review sentences across all entities.
	
	For both ablation setups, we observe almost no change in perplexity, aspect coverage and distinct $n$-grams metrics. 
	
    \begin{table}[t!]
    	\centering
    	\resizebox{0.5\textwidth}{!}{
	\begin{tabular}{c l |c c c} 
		\toprule[1pt]
		{\textsc{Dataset}} & \textsc{Method} & {R1} & {R2} & {RL} \TBstrut\\ 			
		\midrule[1pt]
    	\multirow{2}{*}{\textsc{Space}}& {\ST} & \textbf{42.48} & \textbf{13.48} &  \textbf{26.40} \Tstrut\\
		& \hspace*{0.35cm} w/ Herding &  39.69 & 10.30 & 22.81 \\	
		& \hspace*{0.35cm} w/ Optimal Transport & 38.38 & 9.34 & 22.38 \\
		& \hspace*{0.35cm} w/ Clustering & 30.00 & 4.35 & 17.66 \Bstrut\\	
		\midrule[1pt]
    	\multirow{2}{*}{\textsc{Amazon}}& {\ST} & \textbf{32.03} & \textbf{5.38} & {16.47} \Tstrut\\
		& \hspace*{0.35cm} w/ Herding &  30.36 & 4.95 & 15.67 \\
		& \hspace*{0.35cm} w/ Optimal Transport & 31.45 & 5.23 & \textbf{17.12} \\
		& \hspace*{0.35cm} w/ Clustering & 31.42 & 5.27 & 16.58 \Bstrut\\	
		\bottomrule[1pt]
	\end{tabular}
}
    	\caption{Summarization performance of {\ST} with different sentence selection schemes on \textsc{Space} and \textsc{Amazon} datasets.}
    	\label{tab:herding}
    \end{table}
\end{itemize}

\subsection{Variations of Sentence Selection}
\label{sec:variations}
\noindent(a) {\textbf{Herding} \cite{chen2012super}: In this setup, we modify selection mechanism of {\ST} by updating the mean representation every time a sentence is selected. We consider the mean of the sentences that have not been selected so far. The intuition behind this approach is that the next selected sentence should best capture information, which is not present in the summary so far. The sentence selection process is described below:
	}
	
	\begin{equation}
    	\alpha^s_t = \max_{\alpha^s} \mathcal{R}(\alpha^s) = \max_{\alpha^s} \Delta(\bar{\alpha}_t, \alpha^s)
    	\label{eqn:herding}
    \end{equation}
    \begin{equation}
        \bar{\alpha}_t = \mathop{\mathbb{E}}\limits_{s \sim (S_e \setminus \hat{O}_e)}[\alpha^s]
    \end{equation}
{where $\alpha^s_t$ is the representation selected at time step $t$, $\bar{\alpha}_t$ is mean representation of the set of sentences that are not part of the summary yet and $\hat{O}_e$ is the set of selected sentences so far. Table~\ref{tab:herding} reports the result of this setup. We observe a significant drop in performance compared to {\ST}. We believe that removing the selected sentences skews the mean towards outlier review sentences resulting in a drop in performance. }

\noindent(b) {
\textbf{Optimal Transport}: In this setup, we consider the Wasserstein distance between two probability distributions. Wasserstein distance \cite{peyre2019computational} arising from the concept of optimal transport takes into account the underlying geometry of the representation space. Let $\mathcal{M}^1_+(\mathbb{R}^d)$ be the space of probability distributions defined on $\mathbb{R}^d$ with $d \in \mathbb{Z}^+$. Wasserstein distance between two arbitrary probability distributions $\mu \in \mathcal{M}^1_+(\mathcal{X})$ and $\nu \in \mathcal{M}^1_+(\mathcal{Y})$ is denoted by $\mathcal{W}(\mu, \nu)$. Following \cite{colombo-etal-2021-automatic}, we compute a Wasserstein barycenter of all sentences for each head $h$ as:}
\begin{equation}
    \mu^c_h = \argmin_{\mu \in \mathcal{M}^1_+(\mathbb{R}^d)} \sum_{i=1}^{\lvert S_e\rvert} \mathcal{W}(\mu, \alpha^s_h)
\end{equation}

{
The overall representation for the barycenter is $\mu^c = [\mu^c_1, \ldots, \mu^c_H]$. Next, we derive the relevance score of each sentence $s$ with the barycenter as:}
\begin{equation}
    \mathcal{R}(\alpha^s) = - \sum_{h=1}^{H} \mathcal{W}(\mu^c_h, \alpha^s_h)
    \label{eqn:wasserstien}
\end{equation}
{
As shown in Equation~\ref{eqn:wasserstien}, we select sentences with low Wasserstein distance from the barycenter. We report the results for this optimal transport setup in Table~\ref{tab:herding}. We find that the performance of this setup is significantly lower than {\ST} on \textsc{Space} dataset, but comparable to other baselines on \textsc{Amazon} dataset.}

\noindent(c) {\textbf{Clustering-based Sentence Selection}: In this setup, instead of selecting sentences similar to the mean representation, we identify clusters formed by the representations. For clustering we flatten the sentence representation $\alpha^s \in \mathbb{R}^{HK}$, and use $k$-means\footnote{We experimented with algorithms (like Affinity Propagation, DBSCAN) that identify clusters automatically, but found them to struggle with outliers. K-means performed better than them albeit requiring finetuning of the hyperparameter.} clustering ($K$ is a hyperparameter). We select sentences that are representative samples in each cluster. The relevance score for each sentence is computed as follows:
}
\begin{equation}
	\mathcal{R}(\alpha^s) = - \lvert\lvert \alpha^s - {\alpha}_{\mathcal{C}} \rvert\rvert^2_2 + \gamma \lvert \mathcal{C} \rvert
	\label{eqn:clustering}
\end{equation}

\noindent{
where $\alpha_{\mathcal{C}}$ is the representation of the cluster center where $s$ belongs, and $\lvert\mathcal{C}\rvert$ is the size of the cluster. The first term in Equation~\ref{eqn:clustering} penalizes the relevance of a sentence for being too far away from the cluster center, and the second term selection of samples from a large cluster. The hyperparameters $\gamma = 0.005, K = 5$ in our experiments, were selected using the development set performance. In Table~\ref{tab:herding}, we observe that this clustering-based sentence selection work poorly for \textsc{Space} dataset but the performance on \textsc{Amazon} is decent. The performance on \textsc{Space} dataset is poor as it has a large number of reviews, and identification of representative clusters is difficult using this approach.}

\subsection{Extended Analysis}
\label{sec:extended-analysis}

\begin{table}[t!]
	\centering
	\resizebox{0.43\textwidth}{!}{
	\begin{tabular}{ l l |c c c} 
		\toprule[1pt]
		 \textsc{Dataset} & \textsc{Method} & {R1} & {R2} & {RL} \TBstrut\\ 			
		\midrule[1pt]
    	\multirow{3}{*}{\textsc{Space}} & {\ST} & {42.48} & {13.48} &  {26.40} \Tstrut\\
		& \hspace*{0.55cm}w/o L1 & 41.01 & 11.91 & 24.23  \\
		& \hspace*{0.55cm}w/o H & 38.70 & 10.45 & 22.87 \\
        \midrule[1pt]
    	\multirow{3}{*}{\textsc{Amazon}}& {\ST} & {{32.03}} & {5.38} & {16.47} \Tstrut\\
		& \hspace*{0.55cm}w/o L1 & 29.16 & 4.77 & 16.19  \\
		& \hspace*{0.55cm}w/o H & 29.60 & 4.85 & 16.63  \Bstrut\\
		\bottomrule[1pt]
	\end{tabular}
}
	\caption{ Performance of {\ST} in different configurations of sparsity constraints. }
	\label{tab:sparsity}
\end{table}

\begin{figure*}[t!]
	\centering
	\small
    \input{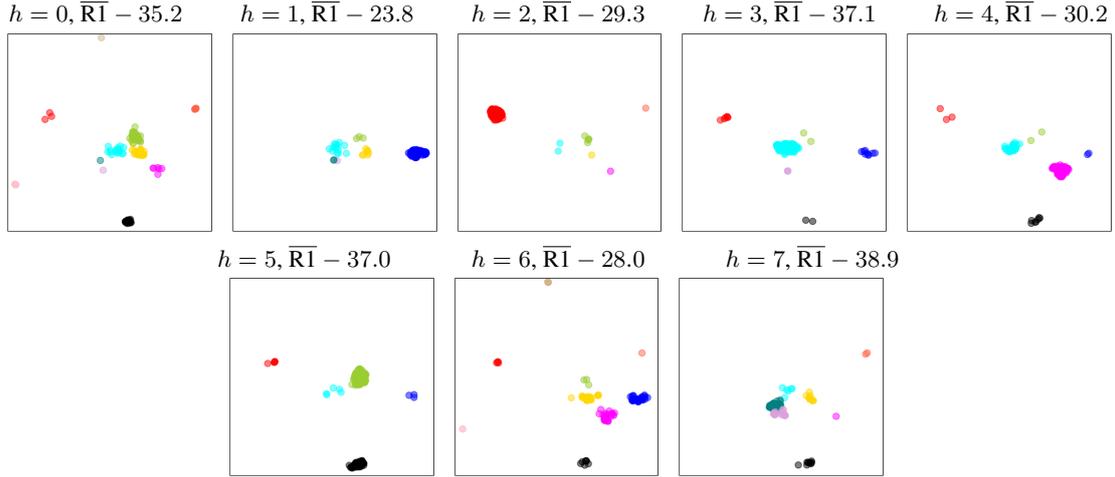}
	\caption{Head-wise visualization of UMAP \cite{mcinnes2018umap}  dictionary element projections.}
	\label{fig:head-dict-viz}
\end{figure*}

\begin{table*}[t!]
	\footnotesize
	\centering
	\begin{tabular}{p{0.24\textwidth} | p{0.23\textwidth} | p{0.21\textwidth} | p{0.2\textwidth}} 
		\toprule[1pt]
		\multicolumn{1}{c|}{\textbf{Human}} & \multicolumn{1}{c|}{\textbf{\ST}} & \multicolumn{1}{c|}{\textbf{QT}} & \multicolumn{1}{c}{\textbf{Copycat}} \TBstrut\\ 			
		\midrule[1pt]
		& & & \\[-0.5em]
		All staff members were friendly, accommodating, and helpful. The hotel and room were very clean. The room had modern charm and was		nicely remodeled. The beds are extremely comfortable. The rooms are	quite with wonderful beach views. The food at Hash, the restaurant in lobby, was fabulous. The location is great, very close to the beach. It’s a	longish walk to Santa Monica. The price is very affordable. 
		& The staff is great.	The Hotel Erwin is a great place to stay.	The staff were friendly and helpful.	The location is perfect.	We ate breakfast at the hotel and it was great.	The hotel itself is in a great location.	The service was wonderful.	It was great.	The rooms are great.	The rooftop bar HIGH was the icing on the cake.	The food and service at the restaurant was awesome.	The service was excellent.
		& Great hotel. We liked our room with an ocean view. The staff were friendly and helpful. There was no balcony. The location is perfect. Our room was very		quiet. I would definitely stay		here again. You’re one block		from the beach. So it must		be good! Filthy hallways.		Unvacuumed room. Pricy,		but well worth it.
		
		&  This hotel is in a great location, just off the beach. The		staff was very friendly and		helpful. We had a room with		a view of the beach and ocean. The only problem was that our room was on the 4th floor with a view of the ocean. If you are looking for a nice place to sleep then this is the	place for you.
		\TBstrut\\
		\bottomrule[1pt]
\end{tabular}
	\caption{ Human-written  and system generated summaries from {\ST}, QT and Copycat. We showcase the summary for the same instance reported by previous works.}
	\label{tab:summaries}
\end{table*}
\begin{table*}[h!]
	\small
	\centering
        \begin{tabular}{@{}p{\textwidth}@{}}
		\toprule[1pt]
		\textbf{Food}: The food and service at the restaurant was awesome.	The food at Hash, the restaurant just off of the lobby, was fabulous for breakfast.	The food was excellent (oatmeal, great wheat toast, freshberries and a tasty corned beef hash). \\
		\midrule
		\textbf{Location}: The Hotel Erwin is a great place to stay.	The hotel is not only in the perfect location for the ideal LA beach experience, but it is extremely hip and comfortable at the same time.\\
		\midrule
		\textbf{Cleanliness}: The room was spacious and had really cool furnishings, and the beds were comfortable.	The room itself was very spacious and had a comfortable bed.	We were upgraded to a partial ocean view suite and the room was clean and comfortable.\\
		\midrule
		\textbf{Service}: The hotel staff were friendly and provided us with great service.	The staff were friendly and helpful.	The staff was extremely helpful and friendly.	The hotel staff was friendly and the room was well kept.\\
		\midrule
		\textbf{Building}: The rooftop bar at the hotel, "High", is amazing. The rooftop bar HIGH was the icing on the cake.	The Hotel Erwin is a great place to stay. The best part of the hotel is the 7th floor rooftop deck.\\
		\midrule
		\textbf{Rooms}: The room was spacious and had really cool furnishings, and the beds were comfortable. The room itself had a retro 70's feel with a comfortable living room and kitchen area, a separate bedroom with a nice king size bed, and a sink area outside the shower/toilet area.\\
		\bottomrule[1pt]
\end{tabular}
	\caption{Aspect-wise summaries generated by {\ST}.}
	\label{tab:asp-summaries}
\end{table*}

\begin{figure*}[t!]
	\centering
    \small 
	Epoch 4 \hspace{2.5cm} 
	Epoch 5 \hspace{2.5cm} 
	Epoch 6 \hspace{2.5cm} 
	Epoch 7 \par
	\includegraphics[width=0.22\textwidth]{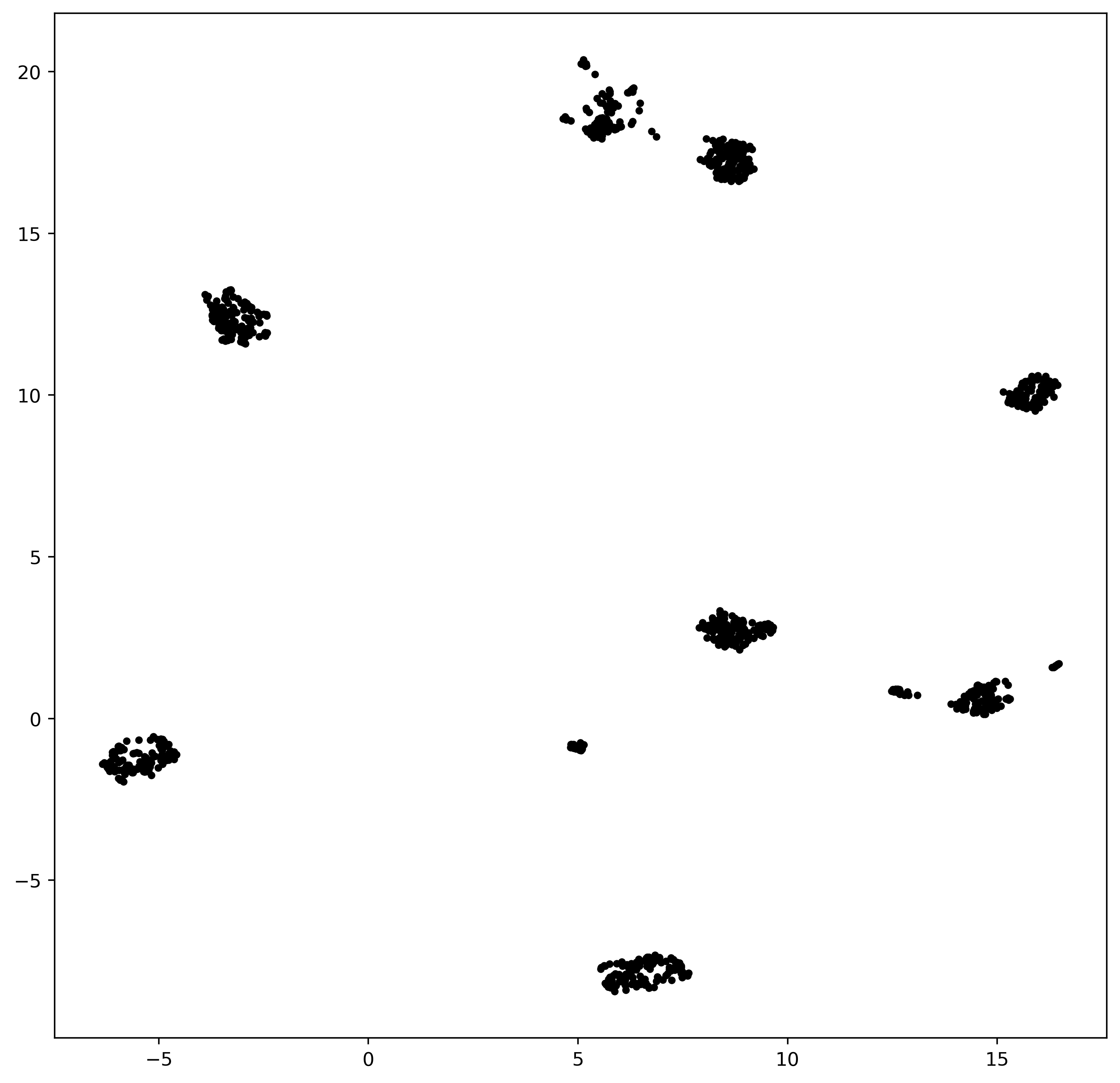}
	\includegraphics[width=0.22\textwidth]{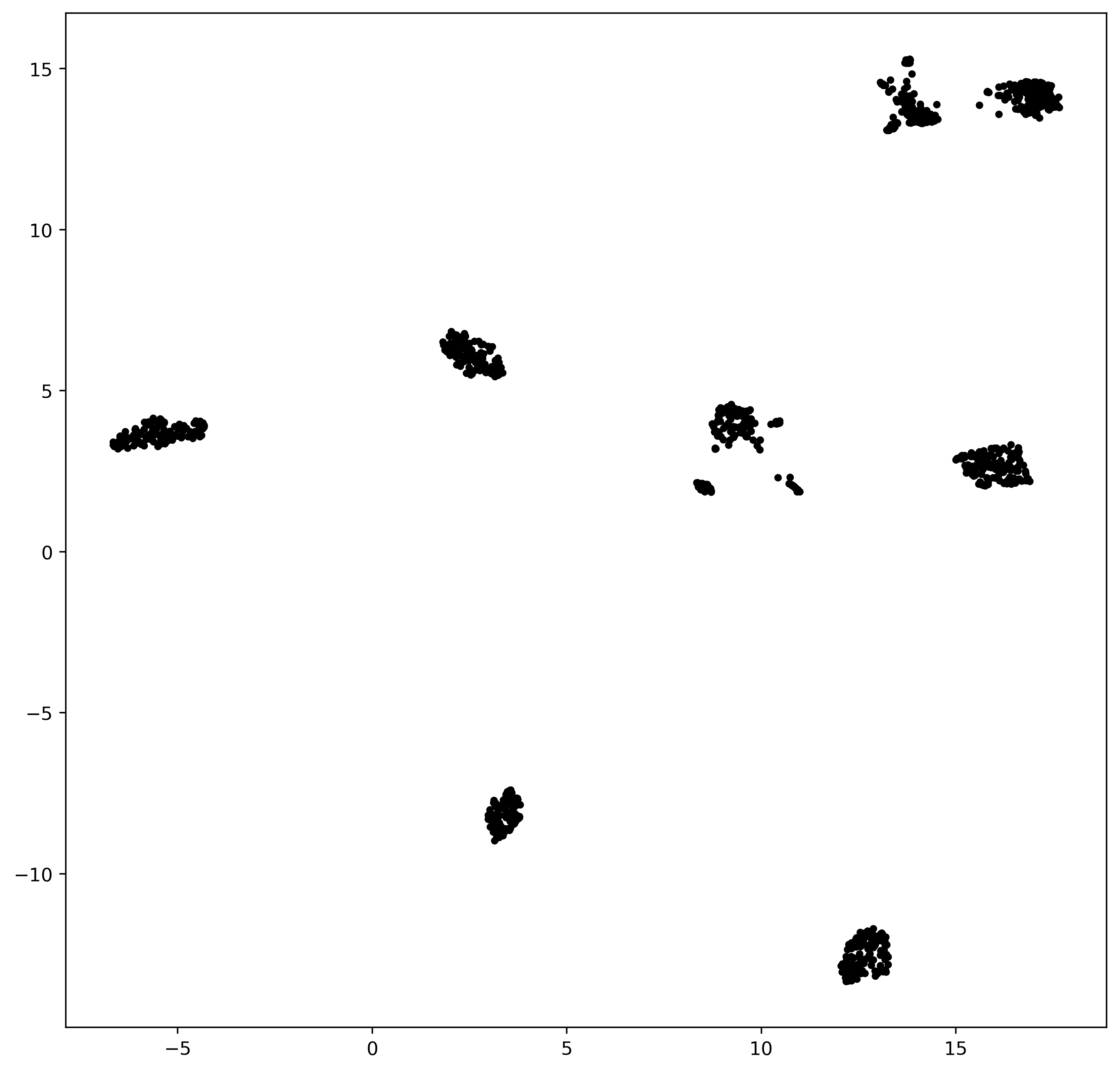}
	\includegraphics[width=0.22\textwidth]{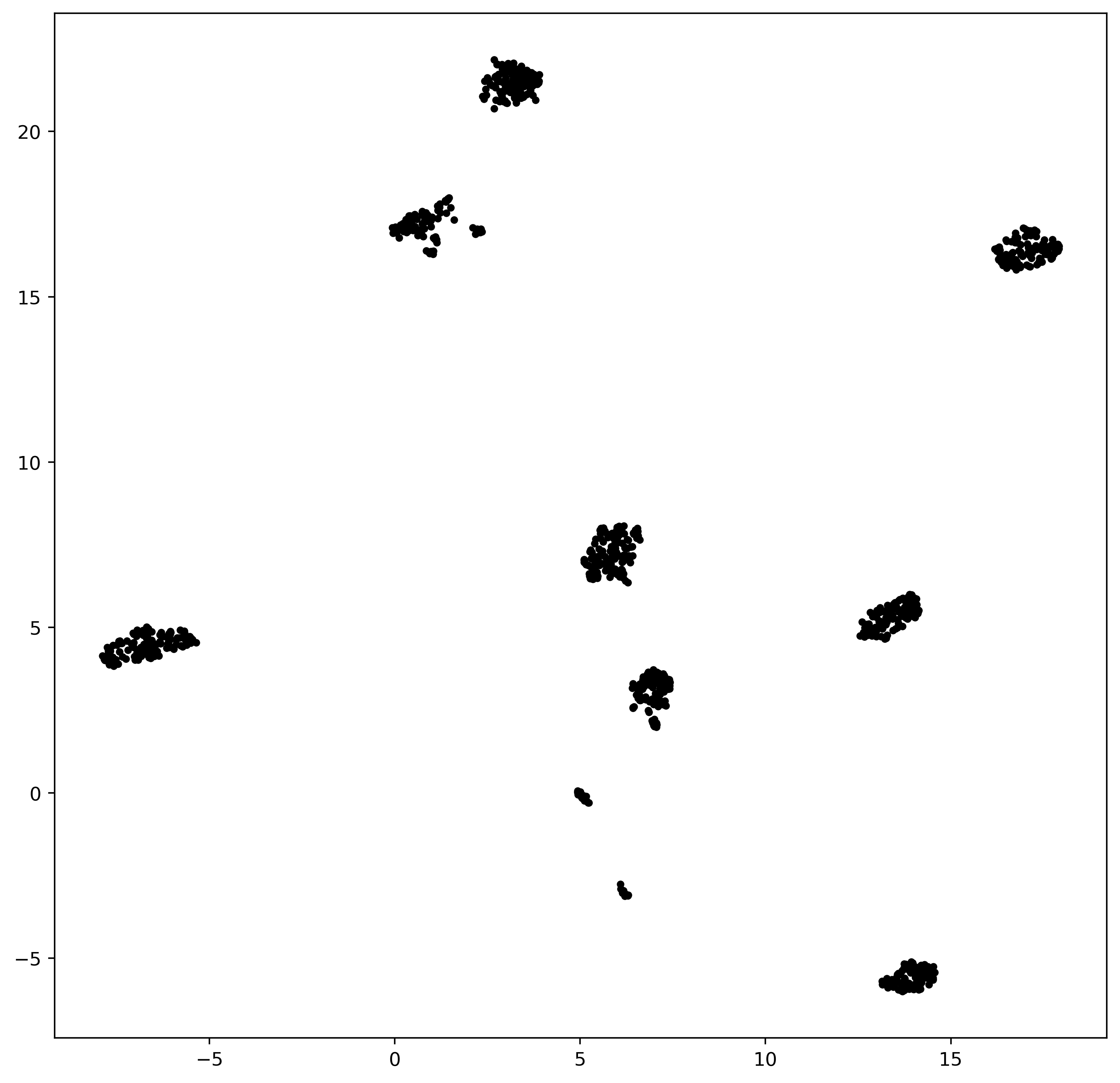}
	\includegraphics[width=0.22\textwidth]{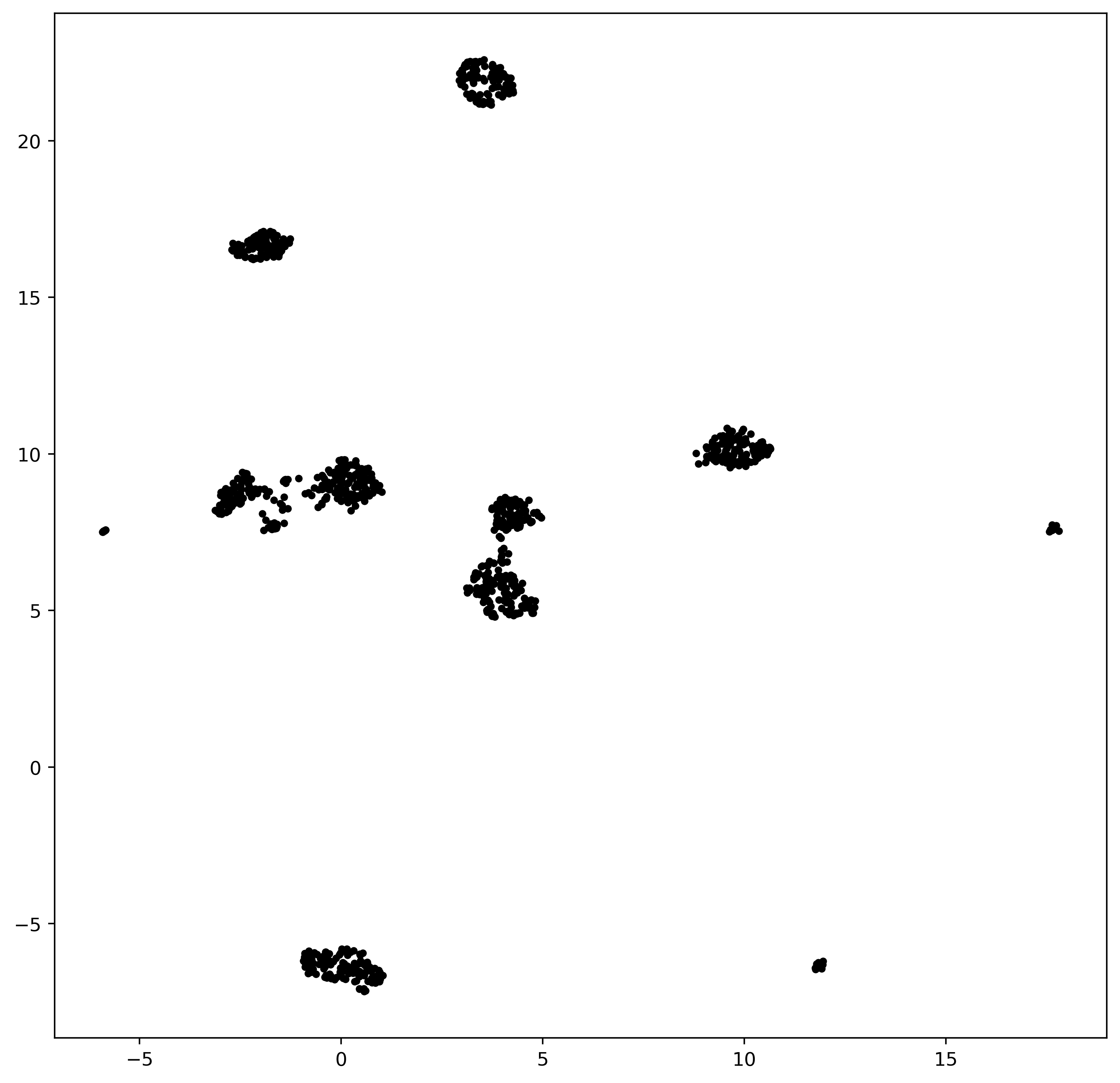}
	
	Epoch 8 \hspace{2.5cm} 
	Epoch 9 \hspace{2.5cm} 
	Epoch 10 \par
	\includegraphics[width=0.22\textwidth]{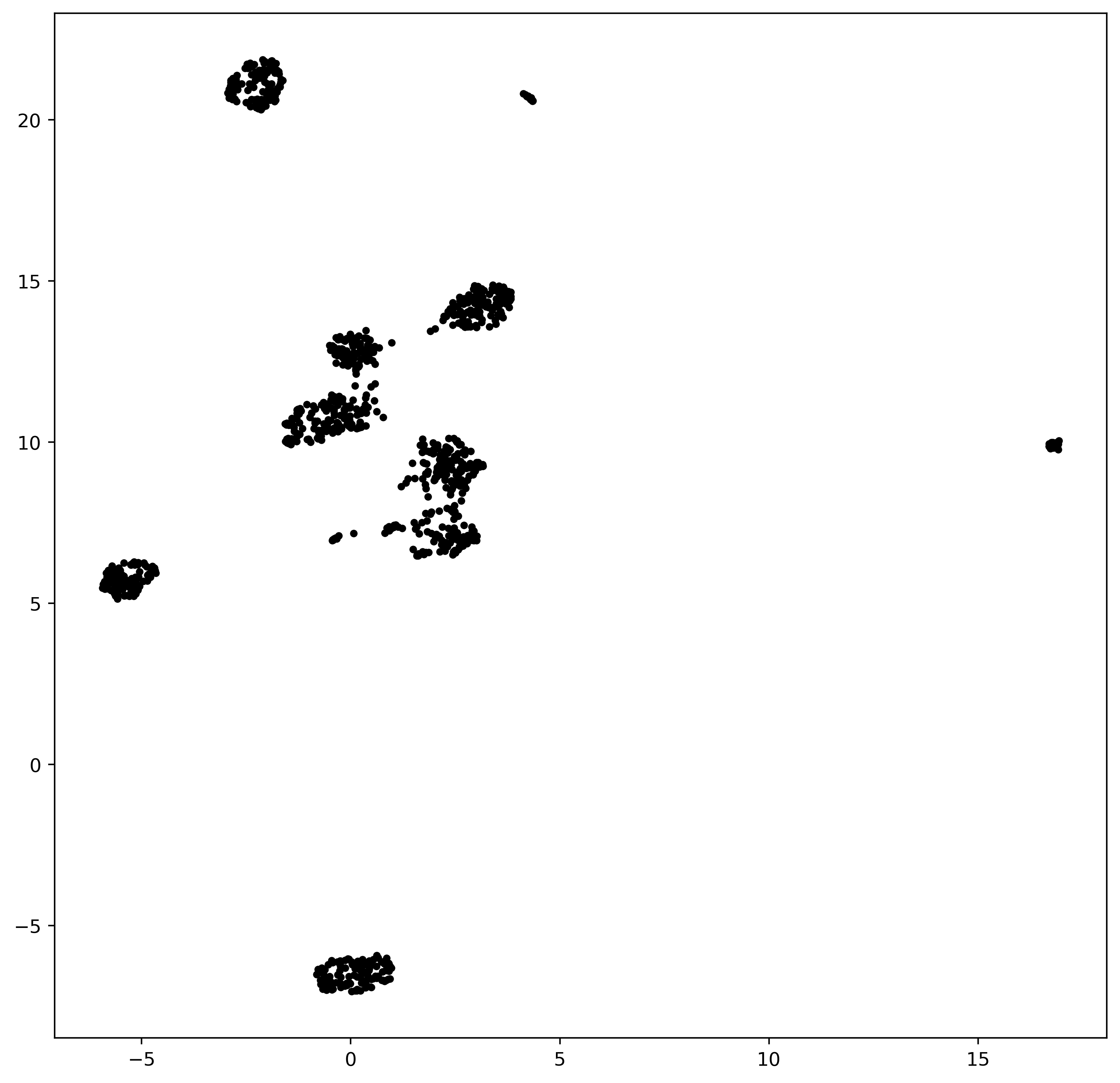}
	\includegraphics[width=0.22\textwidth]{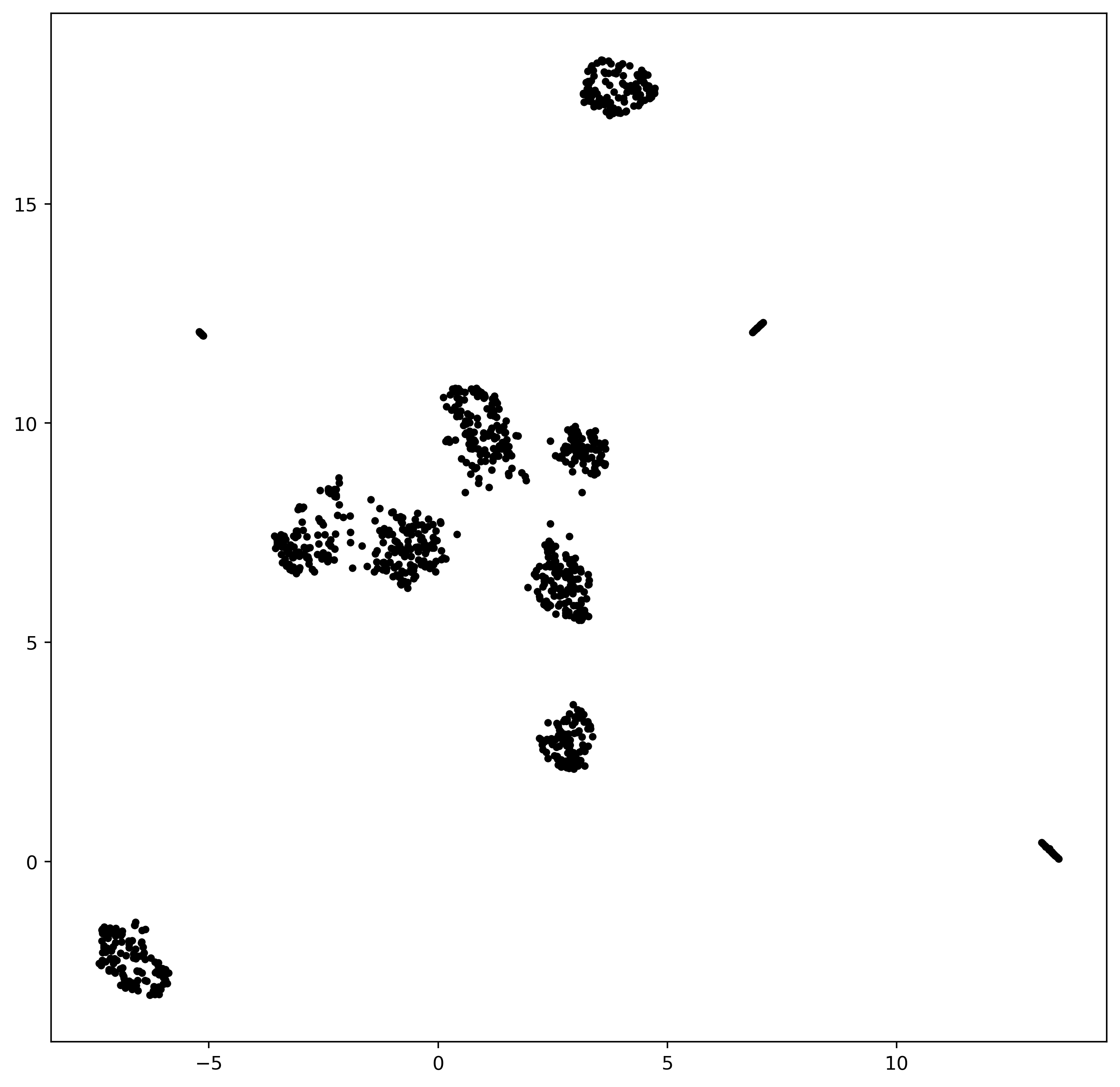}
	\includegraphics[width=0.22\textwidth]{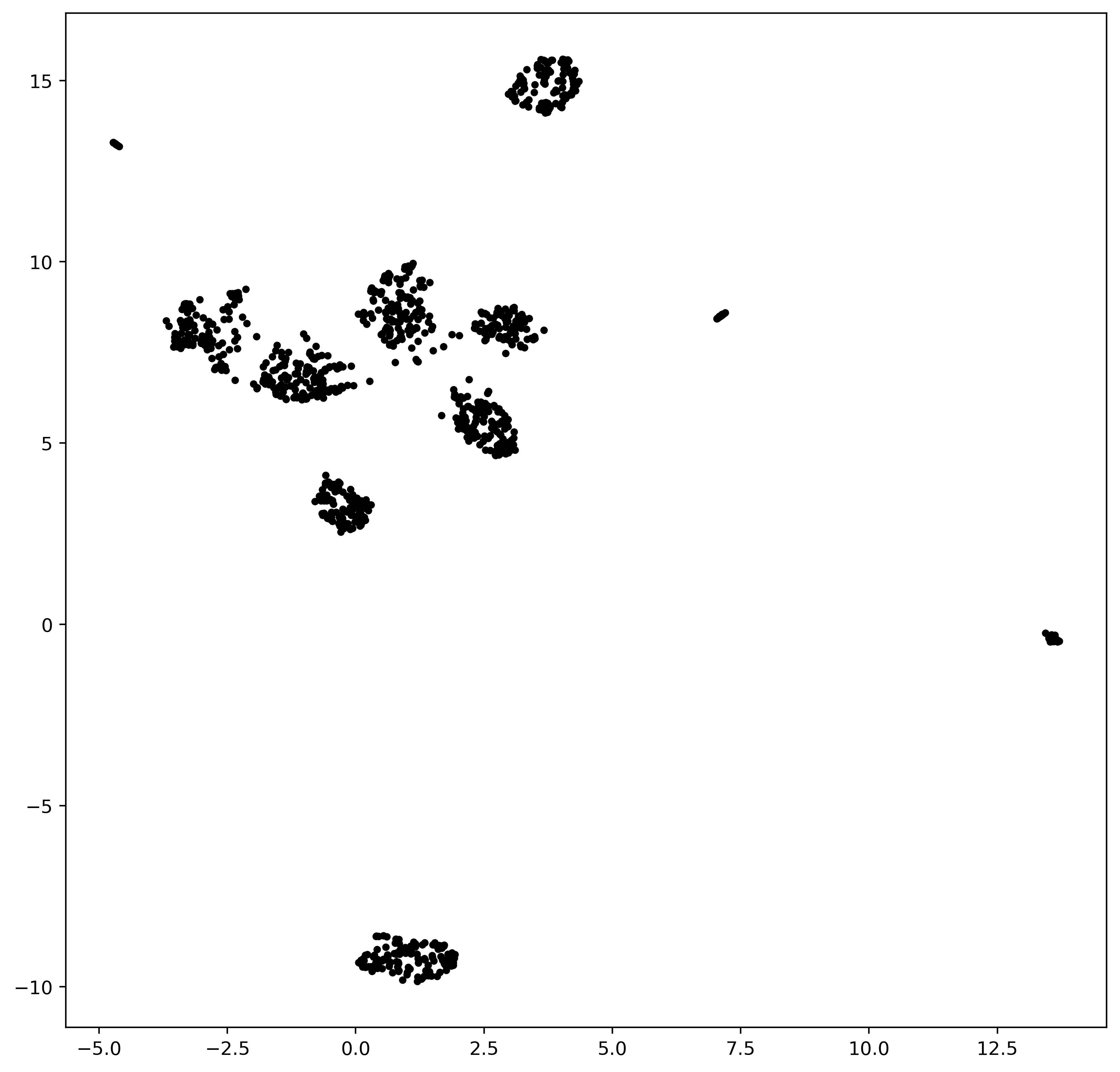}
	\caption{UMAP \cite{mcinnes2018umap}  projections of dictionary element over different epochs (warmup epoch \#4 to epoch 10). We observe that dictionary elements gradually evolve to form clusters over the epochs.}
	\label{fig:dict-evolution}
\end{figure*}

\noindent{(a) \textbf{Efficacy of Sparsity Losses}: In this section, we evaluate the performance of {\ST} in different configurations of sparsity losses. Specifically, we gauge {\ST}' performance when L1-loss and entropy loss are removed. Table~\ref{tab:sparsity} reports the results with different loss setups. We observe a drop in performance when either of the sparsity losses are removed. This shows that ensuring sentence representations are a sparse combination of semantic units helps in summarization.}

\noindent(b) \textbf{Head-wise Analysis}: We analyze whether there is a correlation between the head-wise representations and clusters formed by dictionary elements. For each dictionary element, we compute the average attention ($\alpha_h$) it receives from each head $h$, and assign the element to a head where it received the maximum mean attention (head-wise dictionary elements are shown in Figure~\ref{fig:head-dict-viz}). We also compute the performance of general summarization when only a single head representation is considered $\Delta(\alpha^s, \bar{\alpha}) = \mathrm{KL}(\bar{\alpha}_h, \alpha_h^s)$. In Figure~\ref{fig:head-dict-viz}, we observe that heads that have instances in multiple dictionary element clusters ($h=0, 3, 5, 7$) perform better than heads where instances are concentrated over few clusters ($h=1, 2$).

\noindent(c) \textbf{Output summaries}:
Table~\ref{tab:summaries} shows the summaries generated by {\ST}, QT and Copycat along with human-written summary. We observe that {\ST} selects well formed sentences, avoiding truncated sentences or the ones in a first-person setting.
Table~\ref{tab:asp-summaries} reports the summaries generated by {\ST} for different aspects of a hotel entity. We observe that {\ST} is able to produce summaries that talk about the specific aspect only.  

\noindent(d) \textbf{Evolution of Dictionary Representations}: We plot the UMAP projections of dictionary elements from epochs 4 (after encoder warmup is complete) to 10 in Figure~\ref{fig:dict-evolution}. During the training process, we observe that the UMAP project of dictionary elements form a set of clusters. We observe the first signs of cluster formation in epoch 7, which becomes more distinct over the later epochs.

\begin{table}[t!]
	\centering
	\resizebox{0.35\textwidth}{!}{
	\begin{tabular}{ l|c c c} 
		\toprule[1pt]
		\textsc{Space} [General] & R1 & R2 & RL \TBstrut\\ 			
		\midrule[1pt]
		QT & 36.1 & 7.6 & 20.2 \Tstrut\\
		QT (+SS) & 35.7 & 8.1 & 22.4 \Bstrut\\
		\hline
		{\ST} & \textbf{37.8} & \textbf{9.7} & \textbf{22.8}\TBstrut\\
		\bottomrule[1pt]
	\end{tabular}
}
	\caption{Summarization performance with SemAE's sentence selection (SS) scheme using representations from QT and SemAE. We also report the performance of the baseline QT. The experiments were conducted on 5\% \textsc{Space} dataset.}
	\label{tab:qt-ablation}
\end{table}

\noindent(e) \textbf{Ablations with QT}: In this section, we analyze the efficacy of our sentence selection (SS) module. We evaluate the summarization performance using our sentence selection scheme by retrieving sentence representations from QT and SemAE. The experiments were performed using 5\% data from the \textsc{Space} dataset. For QT's representations, we obtain $\alpha_h$ (Equation~\ref{eqn:alpha_h}) as follows:
\begin{equation}
    \alpha_h = \mathrm{softmax}(-\lvert\lvert s_h - D \rvert\rvert^2_2)
\end{equation}
In Table~\ref{tab:qt-ablation}, we observe that incorporating our sentence selection (SS) improves QT's performance in terms of ROUGE-2 and ROUGE-L scores, with a small drop in ROUGE-1. However, the performance still falls behind SemAE, showcasing that the our representation learning model complements the sentence selection scheme. From these two results, we can conclude that the better performance of SemAE can be attributed to a combination of the two components.  Note that using QT’s sentence selection with SemAE’s representations is not feasible as SemAE doesn’t quantize sentences to a single latent code.



\end{document}